\documentclass[lettersize,journal]{IEEEtran}
\usepackage{amsmath,amsfonts}
\usepackage{algorithmic}
\usepackage{algorithm}
\usepackage{array}
\usepackage{textcomp}
\usepackage{stfloats}
\usepackage{url}
\usepackage{verbatim}
\usepackage{graphicx}
\usepackage{cite}
\usepackage{booktabs}
\usepackage{hyperref}
\usepackage{multirow}
\usepackage{amssymb}
\usepackage{subcaption}
\usepackage{makecell}
\usepackage[table]{xcolor}
\definecolor{darkgreen}{RGB}{0,128,0}
\definecolor{mediumgreen}{RGB}{0,160,0}
\definecolor{darkred}{RGB}{139,0,0}
\newcommand\Rt[1]{\textcolor{black}{#1}}
\newcommand{\pms}[1]{\textsubscript{\(\pm\){#1}}}

\begin{document}

\title{Multi-level Collaborative Distillation Meets  Global Workspace Model: A Unified Framework for OCIL}

\author{Shibin Su, Guoqiang Liang, De Cheng, Shizhou Zhang, Lingyan Ran
        
\thanks{Manuscript received XXX. This work was supported in part by the National Natural Science Foundation of China (No. 62376218, 62101453, 62576262); in part by Guangdong Basic and Applied Basic Research Foundation (No. 2023A1515011298); in part by the Natural Science Basic Research Program of Shaanxi Province (No. 2022JC-DW-08); in part by the Fundamental Research Funds for the Central Universities under Grant QTZX25083. (Corresponding author: Guoqiang Liang; De Cheng.)}

\thanks{Shibin Su, Guoqiang Liang, Shizhou Zhang, and Lingyan Ran are with the School of Computer Science, Northwestern Polytechnical University, Xi’an 710072, China (e-mail: szzhang@nwpu.edu.cn; gqliang@nwpu.edu.cn; lran@nwpu.edu.cn). 

De Cheng is with the School of Telecommunications Engineering, Xidian University, Xi’an 710071, China (e-mail: dcheng@xidian.edu.cn).

}
}

\markboth{Journal of \LaTeX\ Class Files,~Vol.~14, No.~8, August~2021}%
{Shell \MakeLowercase{\textit{et al.}}: A Sample Article Using IEEEtran.cls for IEEE Journals}


\maketitle

\begin{abstract}
Online Class-Incremental Learning (OCIL) enables models to learn continuously from non-i.i.d. data streams. Since samples of the data streams can be seen only once, it is more suitable for real-world scenarios compared to offline learning. However, this constraint intensifies the challenge for OCIL in maintaining an appropriate balance between stability and plasticity. Moreover, under stricter memory buffer constraints in real world, current replay-based methods are less effective. While ensemble methods improve plasticity, they often struggle with stability. Inspired by the Global Workspace Theory (GWT), we propose a novel approach that enhances ensemble learning through a Global Workspace Model (GWM)—a shared, implicit memory that guides the learning of multiple student models. The GWM is formed by fusing the parameters of all students within each training batch, capturing the historical learning trajectory and serving as a dynamic anchor for knowledge consolidation. Like the broadcasting mechanism of GWT, the GWM is redistributed periodically to students, stabilizing learning and promoting cross-task consistency. In addition, we introduce a multi-level collaborative distillation mechanism. It enforces peer-to-peer consistency among students and preserves historical knowledge by aligning each student with the GWM. As a result, student models remain adaptable to new tasks while maintaining previously learned knowledge, striking a better balance between stability and plasticity. Extensive experiments on three standard OCIL benchmarks show that our method delivers significant performance improvement for several OCIL models across various memory budgets. The code is available at \href{https://github.com/susususushi/GWM}{https://github.com/susususushi/GWM}.
\end{abstract}

\begin{IEEEkeywords}
Online Class Incremental Learning, Global Workspace, Knowledge Distillation, Plasticity and Stability Balance
\end{IEEEkeywords}

\section{Introduction}
\label{Section1:Introduction}
Class-Incremental Learning is designed to integrate the knowledge of classes from a stream of data with an evolved distribution \cite{wang2024comprehensive}. Depending on whether the learner has unlimited access to the current task's training data for multiple epochs, existing methods can be divided into two settings: \emph{offline} and \emph{online}. This paper tackles the more challenging \textit{Online Class-Incremental Learning} (OCIL) task, where the model can use data samples for only one epoch of training ~\cite{ghunaim2023real,gu2024summarizing,wu2025dual}. While OCIL is more efficient in terms of memory and computation, the one-epoch training constraint introduces numerous challenges \cite{soutif2023comprehensive}.

To mitigate the notable performance drop of previously learned tasks, known as catastrophic forgetting (CF) \cite{mccloskey1989catastrophic}, most existing OCIL works rely on data replay techniques \cite{wu2025dual,zhou2024balanced, gu2024summarizing,shim2021online,chaudhry2019tiny}. Specifically, a memory buffer is employed to store a few samples from old tasks. Then, an input batch is drawn from the data stream and merged with a randomly selected memory batch for model training. Following this basic framework, several aspects of replay techniques have been explored. Many works focus on designing efficient strategies for memory update \cite{gu2024summarizing,jin2021gradient} or memory retrieval \cite{aljundi2019online,shim2021online}. Besides, some methods explore how to use data stream and memory samples more efficiently such as augmenting classifiers \cite{wang2023cba,lin2023pcr} and developing new loss functions \cite{guo2022online,caccia2022new,liang2024new}.

While these studies have enhanced the overall accuracy through mitigating CF, they neglect the challenge of learning new tasks. 
Due to the one-epoch training constraint, the OCIL model encounters under-fitting and shortcut learning, leading to biased, non-essential features with inadequate generalization ability\cite{wei2023online}. To enhance plasticity, Wang et al. \cite{wang2024improving} first proposed the use of two peer learners to simultaneously learn from data, which is further augmented with a distillation chain. 
While ensemble methods improve plasticity, they often struggle with stability, particularly under much stricter memory constraints in practical applications where replay samples are extremely scarce. A principled mechanism to consolidate and stabilize knowledge across multiple learners remains an open challenge. On the other hand, in cognitive science, a global workspace theory (GWT) has been proposed to model conscious information processing \cite{BAARS200545,DEHAENE2011200,10.1093/acprof:oso/9780195102659.001.1}. It posits that the brain operates via specialized modules that compete for access to a limited-capacity Global Workspace. \Rt{The information ``winning" this competition is broadcast back to all specialists, enabling integration and the formation of stable, accessible memories.} This architecture inherently addresses a dilemma between learning new tasks (specialists) and integrating and broadcasting knowledge (GW). Although this theory has inspired computational models for learning debiased or multi-modal representation \cite{hong2024debiasing,devillers2024semi},  its formal instantiation in Continual Learning, specifically to the online, memory-constrained setting, has not been explored.

In this work, we draw direct inspiration from GWT to improve the ensemble learning framework for OCIL. Technically, we formalize the Global Workspace as a Global Workspace Model (GWM), and specialized cognitive modules as student models. These student models compete to send distinct information to the GWM while the integrated message within the GWM is broadcast back to all students. During the process of competition and broadcast, we treat the model's parameters as an information carrier rather than its intermediate features or output logits. Informed by linear model connectivity \cite{pmlr-v119-frankle20a}, the competitive access is realized through a linear combination of students' parameters. This ensures that the GWM resides in a flat, low-loss region that generalizes well and is robust to perturbations \cite{hochreiter1997flat}. To further preserve a stable knowledge representation over past training batches, the exponential moving average (EMA) is employed to update the GWM, smoothing the optimization trajectory and stabilizing the loss curvature akin to sharpness-aware minimization \cite{foret2020sharpness}. To implement the broadcasting mechanism of GWT, we design a periodic, partial fusion of the GWM’s parameters back into each student. This cyclic injection of globally integrated knowledge actively realigns the students’ optimization paths, preventing deviation from historically stable optima and thereby explicitly enhancing long-term stability. Through these GWT-inspired mechanisms, the GWM serves as a stable reference anchor that continuously guides the students, enabling them to learn new tasks robustly while preserving consolidated knowledge from past tasks.

To efficiently train this interdependent system in a single epoch, we design a multi-level collaborative distillation mechanism. Besides the cross-entropy (CE) loss for each student, we design a local knowledge distillation (KD) loss to align the outputs of the two students. Additionally, we introduce another global KD loss for synchronizing the outputs of GWM and students, thereby guiding the students towards the average trajectory. Through these losses and parameter fusion, the students' parameters are pushed towards the historical optima, mitigating the risk of parameter drift while preserving the model's plasticity. Meanwhile, this effectively flattens the loss basin around the students’ solution, leading to improved stability and generalization across different tasks. The overall framework is a fundamental strategy, which can be applied to a wide range of existing OCIL approaches.

Our main contributions can be summarized as follows:

\begin{itemize}
\item We propose the first algorithmic translation of Global Workspace Theory for OCIL, introducing a formal competition-and-broadcast mechanism to resolve the stability-plasticity dilemma.

\item We devise a multi-level collaborative distillation mechanism to enforce the consistency of students by synchronizing their predictions and to preserve historical knowledge by aligning each student with the GWM.

\item Extensive experiments on three popular OCIL benchmarks demonstrate the effectiveness of the proposed method, achieving new state-of-the-art performance.

\end{itemize}

The rest of this article is organized as follows. Section \ref{Section2:Related Work} gives a brief review of related work. In Section \ref{Section3:Methodology}, we introduce the proposed method. Then we present and analyze the experimental results in Section \ref{Section4}. Finally, Section \ref{Section5:Conclusion} concludes this study and outlines directions for future research.

\section{Related Work}
\label{Section2:Related Work}
\subsection{Online Class Incremental Learning}
\label{Section2.1:Online Class Incremental Learning}
In OCIL, replay-based methods have gained significant attention due to their effectiveness and simplicity \cite{soutif2023comprehensive}. A pioneering replay-based work is Experience Replay (ER), which adopted a reservoir sampling algorithm and a random updating strategy for memory management \cite{chaudhry2019tiny}. Building upon this foundation, numerous replay-based variants have been developed. Some focus on enhancing memory update and retrieval strategies \cite{aljundi2019online, gu2022not, shim2021online}. For instance, MIR \cite{aljundi2019online} retrieved memory samples that are most interfered by the incoming data batch. SSD \cite{gu2024summarizing} condensed stream data into more informative exemplars for efficient storage. 

Meanwhile, other efforts are directed at improving model learning through architectural innovations \cite{wang2022online, wang2023cba, raghavan2024online, wang2024improving, yan2024orchestrate,11196042} and novel optimization objectives \cite{guo2023dealing, gu2022not, caccia2022new, michel2024rethinking, seo2024learning}. Wang et al. \cite{wang2023cba} designed a continual bias adaptor to augment the classifier. To tackle the overfitting-underfitting dilemma, MOSE-MOE \cite{yan2024orchestrate} introduced a stacked sub-experts model, which was optimized by multi-level supervision and reverse self-distillation. Recently, Wu et al. \cite{wu2025dual} introduced a dual-domain division multiplexer to alleviate both inter-task and intra-task bias, which intervenes confounders and multiple causal factors over frequency and spatial domains. 

For the objective function, ER-ACE \cite{caccia2022new} replaced the vanilla CE loss with an asymmetric variant to mitigate representation drift. OCM \cite{guo2022online} maximized mutual information between the old and new representations. \cite{guo2023dealing} introduced a gradient self-adaptive loss to solve the cross-task discrimination problem. UER \cite{lin2023uer} decomposed the conventional logits of the dot product into an angle factor and a norm factor, using the angle component to learn current samples and both components for replay samples. Based on the principle of maximum a posterior estimation, Michel et al. \cite{michel2024learning} devised a novel loss function, enforcing the learned representations to distribute on the unit sphere. Pareto optimization has also been adopted to capture the interrelationship among previously learned tasks \cite{wu2024mitigating}. Zhou et al. \cite{zhou2024balanced} proposed a balanced destruction-reconstruction module, which tries to reduce the degree of maximal destruction of old knowledge to achieve better knowledge reconstruction. Seo et al. \cite{seo2024learning} proposed preparatory data training to induce neural collapse and a residual correction module to reduce discrepancies during inference. To address task recency bias in the combination of the fully connected layer and softmax, supervised contrast learning \cite{khosla2020supervised} has been incorporated into OCIL  \cite{lin2023pcr, mai2021supervised, chen2024exemplar}. For example, SCR \cite{mai2021supervised} combined it and the nearest class mean classifier. PCR \cite{lin2023pcr} further introduced a proxy-based contrastive loss to address the imbalance issue. 

\subsection{Knowledge Distillation in OCIL}
To alleviate CF, KD \cite{wang2021knowledge} has been widely used in offline CL, where the model learned on old tasks serves as a fixed teacher \cite{li2024continual,roy2023subspace,li2022ckdf,lu2024pamk,ji2022complementary}. Conversely, its application in OCIL is still relatively restricted \cite{koh2023online,han2022online}. Because of the only one-epoch training constraint, the model tends to experience issues such as under-fitting and shortcut learning \cite{wei2023online,wu2025dual}, leading to a less effective teacher. As it fails to capture the critical features, KD ultimately becomes more harmful than beneficial for model update \cite{michel2024rethinking}. To address this problem, momentum knowledge distillation was applied to OCIL in \cite{michel2024rethinking}.
Instead of a student and a well-trained teacher, Wang et al. \cite{wang2024improving} applied KD to two peer learners. Although enhancing model plasticity, it lacked an explicit mechanism for mitigating forgetting. In \cite{yan2024orchestrate}, reverse self-distillation was developed to gather knowledge of various experts from shallow to deep. 

Unlike previous research, we focus on addressing the stability challenges of ensemble models without compromising their adaptability. To accomplish this, we develop a GWM and employ multi-level collaborative distillation to steer the learning of various student models.

\section{Methodology}
\label{Section3:Methodology}
\subsection{Problem Definition}
An OCIL model is trained on a non-i.i.d. data stream consisting of $T$ tasks. In the $t$-th task, its training data is $D_t=\{\mathcal{X}^{t}_{n}, y^{t}_{n}\} ^{N_t} _{n=1}$, where $\mathcal{X}^{t}_{n}$ denotes the $n$-th image with label $y^{t}_{n} \in \mathcal{Y}^t$, and $N_t$ is the number of all images in this task. The labels of any two tasks are non-overlapping, that is, $\mathcal{Y}^i \cap \mathcal{Y}^j=\emptyset$ for $\forall \ i, j \in \{1, \ldots ,T \}, i \neq j$. Each sample can be used to train the model only once, unless stored in the fixed-size memory buffer $\mathcal{M}$. During testing, the model is assessed on a test set containing samples of all learned classes.

\begin{figure*}[t]
    \centering
    \includegraphics[width=0.9\linewidth]{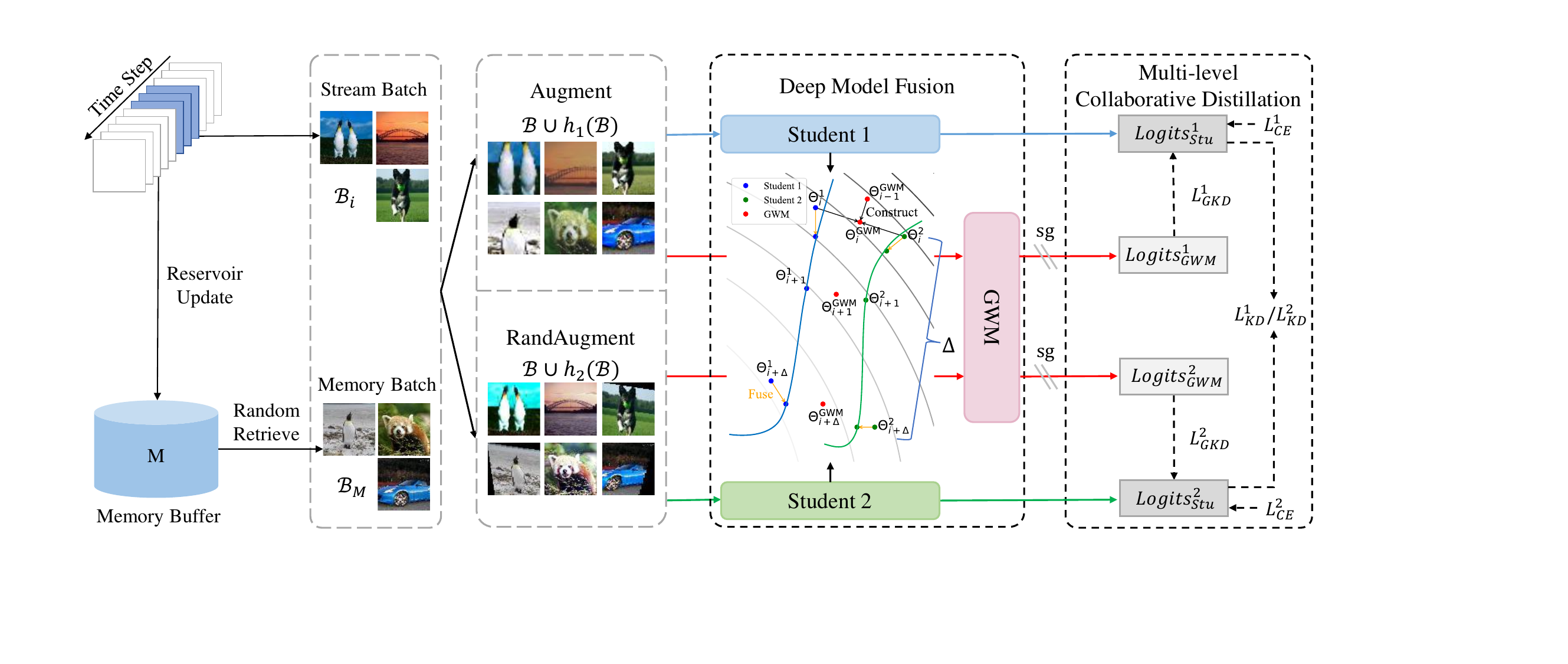}
    \caption{Overview of the proposed method.
    For the combination of a stream data batch and a random retrieved memory batch, we apply two distinct augmentation strategies. The resulting augmented batches and their original batch are fed into two students and Global Workspace Model (GWM) to produce logits. In each training iteration, we use a linear combination of students' parameters to construct the GWM. Moreover, GWM’s parameters are periodically fused back into students. We use blue, dark green and red lines to represent the forward processes of various models. ``sg" denotes the stop-gradient operation.}
    \label{overview}
\end{figure*}

\subsection{Overall Framework}
Drawn inspiration from GWT, we propose to enhance collaborative learning with the introduction of a global workspace model (GWM), aiming to explicitly improve model stability while maintaining plasticity for OCIL. Fig. \ref{overview} illustrates the framework of our method, which consists of two modules: data augmentation and deep model fusion. In data augmentation, each data batch from the current stream is combined with a batch retrieved from the memory buffer, resulting in a batch $\mathcal{B}$. Then, two kinds of data augmentation strategies are applied to the batch $\mathcal{B}$. Next, both the original and the different augmented batches are fed into the deep model fusion module. There are two student networks and a GWM. The two students have identical network structure and the same random initialization. We employ a linear combination of students' parameters as the parameters of the GWM. By further utilizing an EMA with the old GWM, the GWM implicitly captures the historical parameter optimization trajectory, effectively serving as a form of knowledge memory. Through periodically broadcasting the GWM parameters back to all students, the GWM acts as a dynamic anchor to guide student learning.

To optimize our model, we devise a multi-level collaborative distillation mechanism. Besides the traditional CE loss, an online KD loss is employed between the two students. 
By aligning their predictions for distinct augmented versions of identical data, we can enhance the diversity of the two students, thereby improving their plasticity. In addition, we develop another KD loss between the GWM and each of the two students, which compels the students to adhere to historical consensus and strengthens their long-term memory. 

Since there are no constraints on student models, many replay-based CIL methods can serve as the students. In other words, our method can be integrated with many existing replay-based models. In the following, we primarily detail deep model fusion and multi-level collaborative distillation.

\subsection{Deep Model Fusion}
The one epoch training constraint in OCIL poses several challenges, including under-fitting, shortcut learning, and convergence to sharp loss minima. These will result in non-essential features that generalize poorly. Moreover, due to relatively low robustness of model parameters and imbalanced quantity of samples between old tasks and new tasks, optimizing the model for each new task will severely damage the knowledge of old tasks. To address these challenges, existing studies have attempted to combine ensemble learning with online knowledge distillation to enhance plasticity \cite{wang2024improving}. However, previous studies have shown that over-parameterized students tend to diverge from each other during training, even under the supervision of KD loss \cite{neyshabur2020being, singh2020model}. This leads to dispersed parameter space, thereby weakening a unified and robust representation of old knowledge. Besides, their knowledge interaction relies only on the transient state of the current batch, lacking consolidation and integration of historical knowledge. Both of these aspects severely affect the model stability. In summary, for ensemble-based OCIL methods, it is crucial to introduce a stabilization mechanism that coordinates learning modules, integrates, and consolidates historical knowledge while preserving plasticity. This aligns closely with the GWT \cite{BAARS200545,DEHAENE2011200,10.1093/acprof:oso/9780195102659.001.1} in cognitive science. Specifically, GWT achieves system-wide coordination and stability through a global workspace. It selectively receives information from specialists via competition and then broadcasts the integrated knowledge back to the specialists. Based this theoretical inspiration, we develop realization of GWT for OCIL by introducing a formal competition-and-broadcast mechanism to resolve the stability-plasticity dilemma.

To realize the global workspace of GWT, we develop a Global Workspace Model (GWM) by directly merging the parameters of students. Formally, at each training batch $i$, the GWM is built through a competitive linear combination of the student parameters:
\begin{equation}
    \hat\Theta^{\mathrm{GWM}}_{i} = \sum_{m=1}^{M}r^{m}_{i}\Theta^{m}_{i}, \quad \mathbf{r}_i=[r_i^1, \ldots, r_i^M] \sim \text{Dir}(\xi),
\label{eq:GWM}
\end{equation}
where $\Theta_{i}^{m}$ is the parameters of feature extractor for the $m$-th student in the $i$-th training batch, $\hat\Theta^{\mathrm{GWM}}_{i}$ denotes the combined results, $M$ is the number of students, and $\mathbf{r}_i$ is a weight vector constrained by $\sum_{m=1}^M r^m_i=1$. The distribution $\text{Dir}(\xi)$ refers to the Dirichlet distribution parameterized by $\xi \in \mathcal{R}^M$. Since the classifier weight $\mathrm{W}^{\mathrm{GWM}}_{i}$ and $\mathrm{W}^{\mathrm{m}}_{i}$ are updated similarly, we just detail the formulation for the feature extractor for simplicity.
This dynamic aggregation guarantees that the GWM resides in a region of the parameter space interpolated by the two students, approximating a solution that lies inside their shared basin. 
As the GWM represents the surrounding parameter space encompassed by multiple students, its classification loss showcases the maximum and minimum bounds of loss in the vicinity of the student models \cite{Zhang_2023_CVPR}. Thus, optimizing this loss encourages convergence to a flatter, more robust minimum, which implicitly regularizes the overall system and enhances generalization. In this sense, the GWM serves as the concrete computational realization of the Global Workspace, that is a dynamic, aggregated knowledge state formed through competitive access from the students.

In Eq. \eqref{eq:GWM}, the GWM model is updated at each training batch, thus obtaining optimal parameters for the current data batch. However, due to the one-epoch training constraint and extremely limited memory batch size, sample noise and outliers will induce parameter fluctuations, thereby impacting the training process. To establish a stable reference and retain knowledge over more data batches, we integrate the parameters of the current batch with those from the previous batches using the exponential moving average: 
\begin{equation}
    \Theta^{\mathrm{GWM}}_{i} = (1-\alpha)\Theta^{\mathrm{GWM}}_{i-1} + \alpha\hat\Theta^{\mathrm{GWM}}_{i},
\label{eq:GWM_EMA}
\end{equation}
where $\alpha$ is a hyper-parameter, $ \Theta^{\mathrm{GWM}}_{i}$ is the final parameters for the GWM after $i$-th training batch. This fusion not only facilitates a smoother and more stable training procedure but also enables the GWM model to act as a bridge between new and old tasks.  

In GWT, the integrated information within the global workspace is broadcast back to all specialists, which is crucial for maintaining stability. Drawing on this idea, we design a periodic parameter fusion step, which actively injects the consolidated knowledge encoded in the GWM into each students at regular intervals. This can be formulated as:
\begin{equation}
    \Theta^{m}_{i}=(1-\gamma) \Theta^{m}_{i} + \gamma\Theta^{\mathrm{GWM}}_{i}, \text{If } \text{mod}(i,\Delta)=0,
    \label{eq:fusion}
\end{equation}
where $\Delta$ and $\gamma$ are two hyper-parameters denoting the fusion interval and ratio, respectively. $\Delta$ can be set at the task or batch level, such as every task or every fifty batches. Rather than a naive averaging procedure, this operation cyclically broadcasts the historically smoothed optimization trajectory (represented by the GWM) to all students. It explicitly pulls student parameters toward a stable, common region in the parameter space, mitigating excessive divergence caused by training on distinct augmented data. Consequently, while students maintain the plasticity to learn new patterns, their parameters remain anchored by the globally broadcast knowledge, ensuring cohesive and stable learning across tasks.

\subsection{Multi-level Collaborative Distillation}
Our framework includes multiple student models together with a GWM. To promote effective interaction and realize the cooperative-competitive nature of GWT, we design a multi-level collaborative distillation mechanism. It operates on three complementary levels: i) task-specific supervised learning with CE loss to acquire new knowledge, ii) peer-level consistency distillation via KD loss between students to enhance plasticity and robustness, and iii) global-level alignment distillation via KD loss between each student and the GWM to enforce stability and historical knowledge preservation. Through this multi-level design, students can explore diverse solutions while remaining anchored to a unified, stable representation, explicitly tackling the stability–plasticity trade-off in OCIL.

For each student, we first utilize the ground-truth labels of samples to steer its learning process. In particular, if $f(\cdot;\Theta^1)$ represents the feature extractor of student 1, the CE loss of student 1 is computed as follows:
\begin{equation}
    L^1_\text{CE} = -\sum_{(\mathcal{X}, y) \in \mathcal{B}\cup h_1(\mathcal{B})} \log \left( \frac{e^{f(\mathcal{X};\Theta^1) \cdot \textbf{w}^1_y}}{\sum_{j \in \textbf{C}_t} e^{f(\mathcal{X};\Theta^1)\cdot \textbf{w}^1_j}} \right),
    \label{eq:CE}
\end{equation}
where $\textbf{C}_t$ signifies the set of all observed classes up to task $t$, $h_1$ denotes the data augmentation for student 1, $y$ is the ground-truth label of sample $\mathcal{X}$, and $\textbf{w}^1_y$ represents the weight for class $y$ in the classifier of student 1.  

While both students receive distinct augmented data from the same data batch, it is infeasible for them to interact solely through the CE loss. To enhance cross-student interaction, we develop a KD loss to align the predicted probability from the two students. For a sample $\mathcal{X}$, the probability that the student 1 assigns it to class $c \in \textbf{C}_t$ is given by
\begin{equation}
   p^1_c=\frac{e^{f(\mathcal{X};\Theta^1) \cdot \textbf{w}^1_c /\tau}}{\sum_{j \in \textbf{C}_t} e^{f(\mathcal{X};\Theta^1)\cdot \textbf{w}^1_j /\tau}},
   \label{eq:probability}
\end{equation}
where $\tau$ is the temperature hyper-parameter. Therefore, $\textbf{p}^1(\mathcal{X})=\{p^1_c:c\in \textbf{C}_t\}$ symbolizes the softened prediction for all seen classes according to student 1. $\textbf{p}^2(\mathcal{X})$ is derived similarly for student 2. Given these probabilities, the KD loss between students can be expressed as
\begin{multline}
    L^1_\text{KD} = \sum_{(\mathcal{X},y) \in \mathcal{B}} D\left(\textbf{p}^1(\mathcal{X})||\textbf{p}^2(\mathcal{X}) \right)+ \\ D\left(\textbf{p}^1(h_1(\mathcal{X}))||\textbf{p}^2(h_2(\mathcal{X})) \right),
    \label{eq:KD}
\end{multline}
where $D(\cdot \parallel \cdot)$ is the KL divergence. In Eq. \eqref{eq:KD}, we also align the probabilities across various augmented views of the same data, improving the multi-view consistency of the students.

Furthermore, to encourage students towards average trajectory, we introduce another KD loss for outputs of GWM and students. Specifically, for student 1, it can be calculated as  
\begin{equation}
    L^{1}_\mathrm{GKD} = \sum_{(\mathcal{X},y) \in \mathcal{B}\cup h_1(\mathcal{B})} D\left(\textbf{p}^1 (\mathcal{X})||\textbf{p}^\text{GWM}(\mathcal{X}) \right),
    \label{eq:GWM-KD}
\end{equation}
where $\textbf{p}^\text{GWM}(\mathcal{X})$ denotes the predicted probabilities of all observed classes from the GWM model. Utilizing the GWM as a dynamic knowledge anchor, this equation helps avoid significant deviations in students, which might cause the loss landscape to a narrow crevice. By maintaining this, we can retain the model's sensitivity to the crucial features of previous tasks, thus enhancing cross-task generalization.

Ultimately, by combining the above CE loss, KD loss and GKD loss, we can establish the overall multi-level collaborative distillation (MCD) loss for student 1:
\begin{equation}
    L_{\mathrm{MCD}}^1 = L_{\mathrm{CE}}^1+L_{\mathrm{KD}}^1+\lambda L^{1}_\mathrm{GKD},
    \label{eq:okd}
\end{equation} 
where $\lambda$ is a hyper-parameter regulating the proportion of alignment from the GWM model. The loss $L_\text{MCD}^2$ for student 2 can be calculated similarly.

\subsection{Overall Optimization Objective}
To optimize the model parameters, the overall training loss $L^1$ for student 1 incorporates its original loss together with the aforementioned MCD loss,
\begin{equation}
    L^1=L^1_\text{Baseline}+L^1_\text{MCD},
    \label{eq:overall_loss}
\end{equation}
where $L^1_\text{Baseline}$ denotes the loss function of an existing model to which our model adapts. $L^2$ is computed for student 2 in a similar manner.

During inference, a test sample is fed into two student models to produce their respective probabilities. Then, their average is regarded as the final prediction. The entire process of training and inference is detailed in Algorithm \ref{alg:alg1}.

\begin{algorithm}[!t]
\caption{GWM-MCD (Global Workspace Model with Multi-level Collaborative Distillation)}
\label{alg:alg1}
\begin{algorithmic}[1]
\renewcommand{\algorithmicrequire}{\textbf{Input:}}
\REQUIRE Memory buffer size $M_s$; Learning rate $lr$
\renewcommand{\algorithmicrequire}{\textbf{Init:}}
\REQUIRE Memory buffer $\mathcal{M} \leftarrow \{ \} * M_s$; Number of observed samples $n \leftarrow 0$; Parameters of student 1 \{$\Theta^1$,$\textbf{W}^1$\}, student 2 \{$\Theta^2$,$\textbf{W}^2$\} and GWM \{$\Theta^\text{GWM}$, $\textbf{W}^
\text{GWM}$\}; Student 1 and Student 2 are initialized identically; Two AdamW optimizers $optim1$ and $optim2$.

\FOR{$t \in \{1,\ldots,T\}$}
\STATE {\color[RGB]{0, 174, 239} $//$ Training Phase}
\FOR{$\mathcal{B}_i \sim \mathcal{D}_t$}
\STATE $\mathcal{B}_{M} \leftarrow RandomRetrieve(\mathcal{M})$ {\color[RGB]{0, 174, 239} $//$ Memory batch}
\STATE $\mathcal{B} \leftarrow \mathcal{B}_i \cup \mathcal{B}_{M}$ 

\STATE {\color[RGB]{0, 174, 239} $//$ Perform different data augmentation}
\STATE Do data augmentation $\mathcal{\overline{B}}_{1} \leftarrow \mathcal{B} \cup h_{1}(\mathcal{B})$
\STATE Do RandAugment $\mathcal{\overline{B}}_{2} \leftarrow \mathcal{B} \cup h_{2}(\mathcal{B})$


\STATE {\color[RGB]{0, 174, 239} $//$ Update Student 1 parameters. Student 2 is similar} 

\STATE Calculate $L^1_\text{CE}$, $L^1_\text{KD}$, $L^{1}_\text{GKD}$ via Eqs. \eqref{eq:CE}-\eqref{eq:GWM-KD}




\STATE Calculate $L_\text{MCD}^1$ by $L_\text{MCD}^1 \leftarrow L^1_\text{CE}+L^1_\text{KD}+\lambda L^{1}_\text{GKD}$

\STATE Calculate $L^1$ by $L^1 \leftarrow L^1_\text{Baseline}+L^{1}_\text{MCD}$
\STATE $\Theta^1,\textbf{W}^1 \leftarrow optim1(L^1,\Theta^1, \textbf{W}^1, lr)$

\STATE {\color[RGB]{0, 174, 239} $//$ Update GWM parameters}
\STATE Update $\Theta^\text{GWM}$, $\textbf{W}^
\text{GWM}$ for GWM via Eq. (\ref{eq:GWM} \& \ref{eq:GWM_EMA})

\STATE {\color[RGB]{0, 174, 239} $//$ Parameter fusion between GWM and Students}
\IF {$ \mod (i, \Delta) == 0$}
\STATE Update student 1 and 2 using GWM via Eq. \eqref{eq:fusion}
\ENDIF

\STATE {\color[RGB]{0, 174, 239} $//$ Memory Update}
\STATE $\mathcal{M} \leftarrow ReservoirUpdate(\mathcal{M}, \mathcal{B}_i, M_s, n)$
\STATE $n \leftarrow n + |\mathcal{B}_i|$
\ENDFOR

\STATE {\color[RGB]{0, 174, 239} $//$ Inference Phase (Optional)}
\FOR{$ \mathcal{X} \sim \mathcal{D}_{test} $}
\STATE Compute probability for $c \in \textbf{C}_t=\cup_{k=1}^t \mathcal{Y}_k$


\STATE Predict the label by  $y' \leftarrow \arg\max\limits_{ c \in \textbf{C}_t } \frac{ p^1_c(\mathcal{X})+ p^2_c(\mathcal{X})}{2}$

\ENDFOR
\ENDFOR
\end{algorithmic}
\label{alg1}
\end{algorithm}

\section{Experiments}
\label{Section4}

\subsection{Experiment Setup}
\subsubsection{Evaluation Datasets and Metrics}
Following previous work \cite{gu2024summarizing,wang2024improving}, we conduct experiments on three widely used datasets for OCIL: split CIFAR-100~\cite{krizhevsky2009cifar}, split Tiny-ImageNet~\cite{le2015tiny} and split ImageNet-100~\cite{deng2009imagenet}. Both CIFAR-100 and ImageNet-100 are divided into 10 tasks with 10 classes per task. The split Tiny-ImageNet contains 100 disjoint tasks, each of which includes 2 classes. 

Following prior research \cite{wang2024improving, 11196042}, we typically present the final average accuracy (FAA), final relative forgetting (FRF), and average learning accuracy (ALA) to assess performance across all tasks. In contrast to traditional forgetting measure, relative forgetting alleviates the bias towards poor plasticity, providing a fairer evaluation. A superior performance is indicated by a higher FAA or ALA, or a lower FRF. All reported results are the average of 5 runs to reduce randomness. In each experimental run, the order of classes is shuffled for all datasets before splitting the tasks.  

To assess our method under both restricted and standard memory sizes, we perform experiments utilizing extremely limited memory as well as standard memory capacities. Under constrained conditions, the memory size is configured to 0.1K, 0.2K, 0.5K for CIFAR-100, and 0.2K, 0.5K, 1K for Tiny-ImageNet and ImageNet-100. Under standard conditions, the memory size is increased ten times.

\begin{table*}[!t]
\centering
\caption{Comparison of FAA (\%) with constrained memory sizes for individual baselines, those integrated with CCL-DC\cite{wang2024improving}, and those combined with our approach. The best scores are highlighted in \textbf{boldface}. All results are the average of 5 runs.}
    \centering
    \begin{tabular}{cccc|ccc|ccc}
        \toprule
        Dataset & \multicolumn{3}{c}{CIFAR-100 (10 tasks)} & \multicolumn{3}{c}{Tiny-ImageNet (100 tasks)} & \multicolumn{3}{c}{ImageNet-100 (10 tasks)}\\
        \midrule
        Memory Size ($M_s$) & 0.1K & 0.2K & 0.5K & 0.2K &  0.5K &  1K & 0.2K & 0.5K &  1K \\
        \midrule
        ER (2019)\cite{chaudhry2019tiny} & 7.4\pms{0.7} & 8.6\pms{0.5} & 10.9\pms{1.1} & 0.9\pms{0.1} & 0.9\pms{0.1} & 1.0\pms{0.1} & 8.1\pms{1.6} & 11.6\pms{1.6} & 14.9\pms{0.8} \\
        ER+CCL-DC & 11.8\pms{1.1} & 15.1\pms{1.2} & 23.3\pms{1.4} & 3.1\pms{0.5} & 6.8\pms{0.4} & 6.8\pms{1.8} & 11.8\pms{1.3} & 17.5\pms{1.1} & 24.3\pms{0.5} \\
        \rowcolor{blue!10} 
        ER+Ours & \textbf{21.6}\pms{1.4} & \textbf{28.0}\pms{0.9} & \textbf{34.5}\pms{0.8} & \textbf{7.6}\pms{0.5} & \textbf{11.5}\pms{0.7} & \textbf{15.6}\pms{0.8} & \textbf{14.8}\pms{1.1} & \textbf{21.9}\pms{0.3} & \textbf{28.9}\pms{0.7} \\
        \midrule
        SCR (CVPR 2021)\cite{mai2021supervised} & 8.1\pms{0.9} & 11.0\pms{1.1} & 15.3\pms{1.3} & 2.8\pms{0.7} & 6.2\pms{0.3} & 8.6\pms{0.6} & 9.2\pms{0.9} & 14.3\pms{0.5} & 16.9\pms{0.9} \\
        SCR+CCL-DC & 12.0\pms{1.3} & 17.8\pms{1.3} & 28.7\pms{1.2} & 3.1\pms{0.9} & 8.5\pms{0.6} & 13.0\pms{1.0} & 13.2\pms{1.1} & 22.4\pms{1.7} & \textbf{31.9}\pms{1.5}\\
        \rowcolor{blue!10} 
        SCR+Ours & \textbf{12.4}\pms{1.3} & \textbf{19.2}\pms{1.4} & \textbf{29.0}\pms{0.5} & \textbf{3.7}\pms{1.4} & \textbf{9.8}\pms{1.1} & \textbf{14.6}\pms{0.8} & \textbf{13.9}\pms{1.6} & \textbf{22.5}\pms{1.4} & 31.4\pms{1.2} \\
        \midrule
        ER-ACE (ICLR 2022)\cite{caccia2022new} & 10.4\pms{1.4} & 14.1\pms{2.5} & 19.0\pms{2.4} & 3.9\pms{0.6} & 5.9\pms{0.3} & 8.6\pms{0.5} & 13.6\pms{0.8} & 18.2\pms{2.0} & 22.8\pms{1.4} \\
        ER-ACE+CCL-DC & 15.2\pms{0.7} & 19.6\pms{0.6} & 27.4\pms{0.8} & 5.8\pms{0.6} & 8.6\pms{0.7} & 11.5\pms{0.8} & 18.1\pms{1.5} & 26.7\pms{0.4} & 32.5\pms{1.8} \\
        \rowcolor{blue!10} 
        ER-ACE+Ours & \textbf{17.1}\pms{1.3} & \textbf{23.1}\pms{1.3} & \textbf{29.9}\pms{2.3} & \textbf{6.3}\pms{0.5} & \textbf{9.2}\pms{0.5} & \textbf{12.6}\pms{1.0} & \textbf{22.3}\pms{1.3} & \textbf{29.3}\pms{1.6} & \textbf{35.6}\pms{0.8} \\
        \midrule
        OCM (ICML 2022)\cite{guo2022online} & 7.2\pms{0.3} & 10.1\pms{0.9} & 15.5\pms{1.1} & 4.5\pms{0.7} & 7.5\pms{0.4} & 10.1\pms{0.6} & 7.2\pms{1.1} & 9.2\pms{1.2} & 12.2\pms{0.7}\\
        OCM+CCL-DC & 9.4\pms{1.0} & 11.9\pms{0.7} & 17.4\pms{1.6} & 4.7\pms{0.6} & 7.9\pms{0.9} & 12.9\pms{1.1} & 10.4\pms{0.4} & 13.8\pms{1.1} & 19.7\pms{1.8}\\
        \rowcolor{blue!10} 
        OCM+Ours & \textbf{14.9}\pms{0.8} & \textbf{19.9}\pms{0.6} & \textbf{27.4}\pms{0.4} & \textbf{8.1}\pms{1.0} & \textbf{11.9}\pms{1.5} & \textbf{16.1}\pms{0.5} & \textbf{14.4}\pms{3.4} & \textbf{21.3}\pms{1.9} & \textbf{28.3}\pms{2.3}\\
        \midrule
        GSA (CVPR 2023)\cite{guo2023dealing} & 12.1\pms{0.6} & 14.6\pms{0.9} & 19.8\pms{1.8} & 3.5\pms{0.8} & 5.2\pms{0.6} & 7.1\pms{0.6} & 11.5\pms{0.8} & 15.7\pms{1.6} & 20.2\pms{1.1} \\
        GSA+CCL-DC & 12.8\pms{1.2} & 16.5\pms{0.6} & 25.3\pms{0.9} & 2.2\pms{0.5} & 3.8\pms{1.0} & 7.8\pms{2.2} & 12.1\pms{0.9} & 17.8\pms{1.4} & 25.1\pms{1.0} \\
        \rowcolor{blue!10} 
        GSA+Ours & \textbf{18.8}\pms{1.3} & \textbf{24.8}\pms{1.3} & \textbf{32.0}\pms{1.5} & \textbf{7.8}\pms{0.5} & \textbf{11.2}\pms{0.6} & \textbf{14.6}\pms{1.2} & \textbf{16.1}\pms{1.2} & \textbf{23.3}\pms{1.4} & \textbf{30.2}\pms{3.2} \\
        \midrule
        PCR (CVPR 2023)\cite{lin2023pcr} & 15.1\pms{0.7} & 19.0\pms{0.7} & 25.7\pms{1.7} & 5.8\pms{0.8} & 8.3\pms{0.6} & 11.6\pms{0.9} & 16.3\pms{1.9} & 22.1\pms{1.2} & 27.2\pms{1.2} \\
        PCR+CCL-DC & 14.7\pms{1.4} & 19.3\pms{1.2} & 25.5\pms{1.5} & 3.1\pms{1.2} & 8.0\pms{1.0} & 12.6\pms{1.1} & 12.5\pms{1.3} & 18.8\pms{1.5} & 27.1\pms{3.2} \\
        \rowcolor{blue!10} 
        PCR+Ours & \textbf{26.8}\pms{2.2} & \textbf{31.3}\pms{0.5} & \textbf{36.3}\pms{0.9} & \textbf{8.7}\pms{1.0} & \textbf{13.2}\pms{0.7} & \textbf{16.2}\pms{0.5} & \textbf{22.6}\pms{0.7} & \textbf{30.9}\pms{0.4} & \textbf{36.5}\pms{1.2} \\
        \midrule
        MOSE-MOE (CVPR 2024)\cite{yan2024orchestrate} & 14.7\pms{0.7} & 19.4\pms{1.2} & 27.2\pms{0.9} & 4.7\pms{0.5} & 9.1\pms{0.4} & 13.2\pms{1.3} & 16.7\pms{1.8} & 26.0\pms{1.1} & 31.3\pms{2.0} \\
        MOSE-MOE+CCL-DC & 14.4\pms{1.1} & 19.5\pms{2.2} & 31.7\pms{1.1} & 5.1\pms{0.9} & 9.5\pms{1.2} & 16.8\pms{0.6} & 15.8\pms{2.1} & 24.8\pms{2.1} & 36.0\pms{3.1} \\
        \rowcolor{blue!10} 
        MOSE-MOE+Ours & \textbf{20.0}\pms{1.5} & \textbf{27.0}\pms{1.9} & \textbf{36.2}\pms{1.1} & \textbf{8.7}\pms{0.6} & \textbf{14.4}\pms{0.7} & \textbf{19.3}\pms{0.4} & \textbf{18.9}\pms{2.4} & \textbf{29.8}\pms{1.4} & \textbf{39.3}\pms{1.2} \\
        \midrule
        MLG (PR 2025)\cite{liang2025masking} & 15.6\pms{1.1} & 18.6\pms{1.1} & 24.4\pms{0.8} & 3.4\pms{1.1} & 6.0\pms{0.8} & 8.5\pms{1.2} & 17.2\pms{1.6} & 22.8\pms{0.8} & 27.6\pms{1.0}  \\
        MLG+CCL-DC & 16.5\pms{1.2} & 20.9\pms{1.0} & 29.0\pms{1.0} & 5.6\pms{0.9} & 9.8\pms{0.8} & 13.3\pms{1.2} & 20.9\pms{1.8} & 26.7\pms{2.0} & 31.8\pms{1.3}  \\
        \rowcolor{blue!10} 
        MLG+Ours & \textbf{22.7}\pms{0.8} & \textbf{28.3}\pms{1.2} & \textbf{34.6}\pms{0.5} & \textbf{7.5}\pms{0.6} & \textbf{11.2}\pms{0.7} & \textbf{15.0}\pms{1.1} & \textbf{22.8}\pms{2.5} & \textbf{30.9}\pms{1.1} & \textbf{35.4}\pms{0.7} \\
        \midrule
        HPCR (TNNLS 2025)\cite{10839221} & 16.5\pms{0.6} & 21.3\pms{1.1} & 28.2\pms{1.8} & 5.3\pms{0.4} &  8.6\pms{1.0} & 12.4\pms{0.4} & 16.2\pms{1.5} & 21.9\pms{1.3} & 28.0\pms{1.3}\\
        HPCR+CCL-DC & 17.0\pms{2.1} & 22.1\pms{2.0} & 30.6\pms{1.5} & 5.8\pms{1.0} & 10.4\pms{1.3} & 16.0\pms{0.9} & 17.3\pms{1.3} & 26.9\pms{0.5} & 35.1\pms{0.9}\\
        \rowcolor{blue!10} 
        HPCR+Ours &\textbf{27.0}\pms{1.6}&\textbf{31.2}\pms{1.0}&\textbf{37.1}\pms{0.4}&\textbf{9.8}\pms{0.7}&\textbf{13.9}\pms{0.9}&\textbf{17.1}\pms{0.7}&\textbf{26.0}\pms{1.4}&\textbf{33.3}\pms{0.7}&\textbf{39.4}\pms{1.2}\\

        \midrule
        EMI (TMM 2026)\cite{11360301} & 6.8\pms{4.0} & 12.8\pms{1.9} & 23.4\pms{1.0} & 3.7\pms{0.7} & 6.5\pms{0.7} & 11.1\pms{3.5} & 11.9\pms{1.1} & 19.3\pms{3.9} & 29.5\pms{3.2}\\
        EMI+CCL-DC & 12.4\pms{0.9} & 16.4\pms{1.0} & 26.0\pms{1.5} & 3.4\pms{0.5} & 6.2\pms{0.7} & 11.9\pms{1.5} & 13.7\pms{1.5} & 20.5\pms{2.1} & 29.6\pms{2.2}\\
        \rowcolor{blue!10} 
        EMI+Ours & \textbf{20.2}\pms{1.1} & \textbf{26.5}\pms{1.4} & \textbf{35.1}\pms{0.5} & \textbf{6.2}\pms{2.3} & \textbf{14.1}\pms{4.1} & \textbf{20.7}\pms{1.7} & \textbf{21.2}\pms{2.1} & \textbf{27.3}\pms{2.5} & \textbf{34.8}\pms{2.7}\\
        
        \midrule
        ER-DCBA (TPAMI 2026)\cite{11433804} & 10.1\pms{0.9} & 12.4\pms{0.6} & 18.4\pms{1.6} & 1.3\pms{0.1} & 1.1\pms{0.1} & 1.5\pms{0.2} & 13.4\pms{1.3} & 18.6\pms{1.6} & 23.8\pms{1.6}\\
        ER-DCBA+CCL-DC & 13.6\pms{1.0} & 18.0\pms{1.6} & 26.9\pms{2.1} & 4.6\pms{0.3} & 7.6\pms{1.1} & 7.6\pms{3.0} & 13.4\pms{1.8} & 20.7\pms{1.9} & 29.6\pms{1.8}\\
        \rowcolor{blue!10} 
        ER-DCBA+Ours & \textbf{22.3}\pms{1.7} & \textbf{27.5}\pms{1.2} & \textbf{34.8}\pms{0.7} & \textbf{8.2}\pms{0.6} & \textbf{12.9}\pms{0.2} & \textbf{16.6}\pms{0.9} &\textbf{20.6}\pms{1.1} & \textbf{30.7}\pms{2.1} & \textbf{38.1}\pms{0.8}\\

        \bottomrule
    \end{tabular}

\label{restrict memory}
\end{table*}

\begin{table*}[!t]
\centering
\caption{Comparison of FAA (\%) with standard memory sizes for individual baselines, those integrated with CCL-DC\cite{wang2024improving}, and those combined with our approach. The best scores are indicated in \textbf{boldface}. All results are the average of 5 runs.}
    \centering
        \begin{tabular}{cccc|ccc|ccc}
        \toprule
        Dataset & \multicolumn{3}{c}{CIFAR-100 (10 tasks)} & \multicolumn{3}{c}{Tiny-ImageNet (100 tasks)} & \multicolumn{3}{c}{ImageNet-100 (10 tasks)}\\
        \midrule
        Memory Size ($M_s$) & 1K & 2K & 5K & 2K &  5K &  10K & 2K & 5K &  10K \\
        \midrule
        ER (2019)\cite{chaudhry2019tiny} & 14.3\pms{1.0} & 20.3\pms{0.8} & 29.2\pms{1.6} & 0.9\pms{0.2} & 1.3\pms{0.3} & 1.5\pms{0.4} & 14.6\pms{2.0} & 20.9\pms{2.2} & 24.5\pms{2.0} \\
        ER+CCL-DC & 27.9\pms{1.1} & 30.6\pms{2.0} & 31.2\pms{1.4} & 7.2\pms{1.7} & 7.7\pms{0.5} & 8.3\pms{1.9} & 32.1\pms{1.7} & 37.4\pms{1.1} & 40.1\pms{0.4}\\
        \rowcolor{blue!10} 
        ER+Ours & \textbf{38.4}\pms{0.9} & \textbf{42.0}\pms{1.2} & \textbf{43.7}\pms{1.6} & \textbf{19.1}\pms{0.9} & \textbf{19.5}\pms{1.9} & \textbf{18.8}\pms{1.5} & \textbf{32.9}\pms{2.3} & \textbf{40.6}\pms{1.0} & \textbf{42.4}\pms{1.3}\\
        \midrule
        SCR (CVPR 2021)\cite{mai2021supervised} & 18.4\pms{0.8} & 20.3\pms{0.9} & 22.4\pms{0.4} & 10.8\pms{0.8} & 12.7\pms{0.7} & 13.8\pms{0.8} & 19.0\pms{1.4} & 20.4\pms{1.1} & 20.7\pms{1.1}\\
        SCR+CCL-DC & 35.3\pms{0.8} & 38.7\pms{1.0} & 40.3\pms{0.6} & 15.0\pms{0.9} & 15.9\pms{1.2} & 17.1\pms{1.2} & 39.5\pms{1.3} & 44.4\pms{0.9} & 45.1\pms{1.6}\\
        \rowcolor{blue!10} 
        SCR+Ours & \textbf{37.0}\pms{1.0} & \textbf{41.2}\pms{1.1} & \textbf{43.2}\pms{1.2} & \textbf{20.5}\pms{0.8} & \textbf{23.1}\pms{0.6} & \textbf{23.9}\pms{0.8} & \textbf{39.5}\pms{0.9} & \textbf{44.9}\pms{0.7} & \textbf{46.5}\pms{1.3} \\
        \midrule
        ER-ACE (ICLR 2022)\cite{caccia2022new} & 22.3\pms{2.3} & 24.5\pms{0.9} & 26.1\pms{1.4} & 11.4\pms{0.5} & 11.7\pms{0.9} & 10.9\pms{1.1} & 25.9\pms{1.6} & 29.8\pms{1.0} & 32.9\pms{1.0}\\
        ER-ACE+CCL-DC & 31.3\pms{1.0} & 34.7\pms{1.2} & 35.4\pms{1.0} & 14.8\pms{1.1} & 16.6\pms{1.9} & 17.2\pms{2.0} & 38.0\pms{1.6} & 43.0\pms{0.7} & 43.5\pms{2.7}\\
        \rowcolor{blue!10} 
        ER-ACE+Ours & \textbf{34.0}\pms{0.5} & \textbf{37.6}\pms{1.9} & \textbf{39.9}\pms{1.3} & \textbf{15.9}\pms{1.1} & \textbf{19.2}\pms{2.0} & \textbf{19.6}\pms{2.1} & \textbf{39.8}\pms{0.7} & \textbf{44.0}\pms{1.1} & \textbf{46.1}\pms{1.1}\\
        \midrule
        OCM (ICML 2022)\cite{guo2022online} & 16.8\pms{1.6} & 17.6\pms{2.9} & 19.3\pms{1.8} & 12.9\pms{0.8} & 13.3\pms{1.6} & 14.5\pms{1.4} & 14.2\pms{1.6} & 14.6\pms{2.4} & 14.9\pms{2.0}\\
        OCM+CCL-DC & 20.2\pms{2.7} & 21.3\pms{1.5} & 20.5\pms{2.7} & 17.5\pms{1.1} & 19.2\pms{3.0} & 21.1\pms{0.6} & 24.9\pms{1.5} & 30.2\pms{2.6} & 32.6\pms{2.3}\\
        \rowcolor{blue!10} 
        OCM+Ours & \textbf{33.7}\pms{1.2} & \textbf{36.0}\pms{0.8} & \textbf{36.5}\pms{1.0} & \textbf{21.3}\pms{1.4} & \textbf{23.9}\pms{1.1} & \textbf{23.9}\pms{2.0} & \textbf{34.4}\pms{1.8} & \textbf{42.1}\pms{1.2} & \textbf{45.1}\pms{1.6}\\
        \midrule
        GSA (CVPR 2023)\cite{guo2023dealing} & 23.8\pms{0.9} & 26.2\pms{2.2} & 27.9\pms{1.9} & 9.3\pms{1.6} & 12.6\pms{0.8} & 13.7\pms{1.5} & 25.7\pms{1.9} & 32.8\pms{2.1} & 35.9\pms{1.2}\\
        GSA+CCL-DC & 31.5\pms{0.9} & 37.8\pms{1.1} & 41.5\pms{2.5} & 13.5\pms{1.4} & 18.8\pms{1.4} & 18.8\pms{1.1} & 32.7\pms{1.6} & 43.5\pms{0.7} & 48.4\pms{1.8}\\
        \rowcolor{blue!10} 
        GSA+Ours & \textbf{37.7}\pms{0.9} & \textbf{42.7}\pms{1.4} & \textbf{45.8}\pms{0.8} & \textbf{17.9}\pms{1.2} & \textbf{21.0}\pms{0.9} & \textbf{24.5}\pms{1.3} & \textbf{38.4}\pms{1.3} & \textbf{45.9}\pms{1.4} & \textbf{49.0}\pms{1.3}\\
        \midrule
        PCR (CVPR 2023)\cite{lin2023pcr} & 29.3\pms{1.1} & 31.7\pms{1.2} & 33.6\pms{0.9} & 12.8\pms{0.8} & 15.0\pms{1.3} & 15.2\pms{1.2} & 32.7\pms{1.5} & 36.9\pms{2.8} & 38.8\pms{1.8} \\
        PCR+CCL-DC & 30.8\pms{1.5} & 34.1\pms{1.9} & 36.1\pms{1.8} & 13.7\pms{1.3} & 15.6\pms{1.5} & 16.3\pms{1.0} & 34.1\pms{2.0} & 39.6\pms{3.4} & 42.6\pms{2.7}\\
        \rowcolor{blue!10} 
        PCR+Ours & \textbf{40.0}\pms{1.1} & \textbf{42.0}\pms{0.9} & \textbf{43.7}\pms{0.8} & \textbf{19.0}\pms{0.9} & \textbf{21.2}\pms{0.6} & \textbf{22.4}\pms{0.7} & \textbf{41.7}\pms{0.9} & \textbf{45.5}\pms{0.8} & \textbf{46.4}\pms{1.7}\\
        \midrule
        MOSE-MOE (CVPR 2024)\cite{yan2024orchestrate} & 33.3\pms{0.8} & 38.7\pms{0.7} & 41.5\pms{0.8} & 17.3\pms{1.1} & 20.7\pms{1.0} & 20.7\pms{3.4} & 37.3\pms{2.9} & 43.3\pms{2.8} & 43.7\pms{2.5}\\
        MOSE-MOE+CCL-DC & 40.6\pms{1.1} & 46.8\pms{1.1} & 50.3\pms{0.6} & 22.1\pms{0.5} & 27.4\pms{2.5} & 26.0\pms{1.7} & 45.8\pms{1.9} & 53.3\pms{0.7} & 55.1\pms{0.6}\\
        \rowcolor{blue!10} 
        MOSE-MOE+Ours & \textbf{42.2}\pms{1.4} & \textbf{48.4}\pms{0.9} & \textbf{51.6}\pms{0.4} & \textbf{24.4}\pms{0.7} & \textbf{29.6}\pms{0.9} & \textbf{31.2}\pms{0.5} & \textbf{46.3}\pms{2.1} & \textbf{53.9}\pms{1.5} & \textbf{56.8}\pms{0.7}\\
        \midrule
        MLG (PR 2025)\cite{liang2025masking} & 28.5\pms{1.0} & 31.3\pms{0.6} & 33.6\pms{0.8} & 10.3\pms{0.7} & 11.7\pms{0.6} & 11.9\pms{0.8} & 31.0\pms{0.8} & 33.3\pms{1.6} & 34.6\pms{0.8}  \\
        MLG+CCL-DC & 32.5\pms{1.2} & 35.2\pms{1.0} & 37.2\pms{0.4} & 15.9\pms{1.1} & 18.4\pms{0.9} & 18.9\pms{1.2} & 34.8\pms{1.2} & 38.1\pms{1.0} & 39.2\pms{1.7}  \\
        \rowcolor{blue!10} 
        MLG+Ours & \textbf{38.4}\pms{1.1} & \textbf{40.8}\pms{0.8} & \textbf{43.5}\pms{0.9} & \textbf{21.0}\pms{1.6} & \textbf{25.1}\pms{0.5} & \textbf{26.4}\pms{0.5} & \textbf{40.0}\pms{1.0} & \textbf{43.5}\pms{1.0} & \textbf{44.5}\pms{0.9} \\
        \midrule
        HPCR (TNNLS 2025)\cite{10839221} & 32.2\pms{0.9} & 35.1\pms{1.1} & 36.4\pms{0.7} & 14.0\pms{1.9} & 16.6\pms{0.8} & 17.7\pms{1.3} & 33.8\pms{0.9} & 39.0\pms{1.5} & 39.9\pms{1.0}\\
        HPCR+CCL-DC & 34.5\pms{4.8} & 38.2\pms{1.9} & 39.6\pms{2.0} & 19.1\pms{1.3} & 21.1\pms{1.3} & 21.8\pms{0.9} & 40.8\pms{1.7} & 44.7\pms{1.4} & 45.1\pms{1.0}\\
        \rowcolor{blue!10} 
        HPCR+Ours &\textbf{41.0}\pms{0.8}&\textbf{43.4}\pms{0.8}&\textbf{45.4}\pms{0.7}&\textbf{20.1}\pms{0.6}&\textbf{22.5}\pms{0.8}&\textbf{24.1}\pms{0.8}&\textbf{43.5}\pms{0.4}&\textbf{46.6}\pms{0.5}&\textbf{48.4}\pms{0.7}\\
        \midrule
        EMI (TMM 2026)\cite{11360301} & 30.9\pms{1.9} & 36.8\pms{0.6} & 41.2\pms{0.8} & 18.4\pms{2.3} & 24.1\pms{1.0} & 25.5\pms{1.5} & 36.9\pms{1.9} & 44.5\pms{2.1} & 47.1\pms{1.5}\\
        EMI+CCL-DC & 33.8\pms{1.8} & 41.5\pms{0.6} & 47.3\pms{1.4} & 18.3\pms{2.0} & 24.1\pms{3.6} & 25.2\pms{3.0} & 40.1\pms{3.2} & \textbf{49.2}\pms{0.9} & 50.6\pms{3.7}\\
        \rowcolor{blue!10} 
        EMI+Ours & \textbf{41.5}\pms{1.0} & \textbf{47.1}\pms{1.1} & \textbf{50.4}\pms{1.2} & \textbf{26.3}\pms{1.4} & \textbf{31.3}\pms{2.9} & \textbf{32.4}\pms{2.4} & \textbf{42.5}\pms{3.4} & 48.4\pms{2.5} & \textbf{50.9}\pms{2.1}\\
        
        \midrule
        ER-DCBA (TPAMI 2026)\cite{11433804} & 23.3\pms{1.3} & 28.9\pms{0.7} & 35.4\pms{0.9} & 1.5\pms{0.4} & 2.0\pms{0.4} & 1.8\pms{0.3} & 25.9\pms{2.0} & 32.4\pms{3.1} & 31.8\pms{1.5}\\
        ER-DCBA+CCL-DC & 32.0\pms{0.4} & 34.5\pms{1.7} & 34.2\pms{1.1} & 8.9\pms{3.9} & 8.7\pms{2.9} & 9.1\pms{2.9} & 36.3\pms{2.3} & 40.9\pms{2.0} & 41.4\pms{0.4}\\
        \rowcolor{blue!10} 
        ER-DCBA+Ours & \textbf{40.2}\pms{0.9} & \textbf{43.6}\pms{0.3} & \textbf{46.4}\pms{0.3} & \textbf{20.2}\pms{3.3} & \textbf{20.8}\pms{1.0} & \textbf{20.9}\pms{1.4} & \textbf{44.3}\pms{1.2} & \textbf{49.4}\pms{1.1} & \textbf{50.5}\pms{2.1}\\        
        \bottomrule
    \end{tabular}

\label{standard memory}
\end{table*}
\subsubsection{Implementation Details}
We employ the full ResNet-18 network without pre-training as the backbone for all methods and datasets. Like \cite{chaudhry2019tiny}, random retrieval and reservoir sampling are used for memory management, which are denoted as $RandomRetrieve(\cdot)$ and $ReservoirUpdate(\cdot)$ in Algorithm \ref{alg:alg1}. As we focus on extremely constrained memory and small memory batch size, the batch size is set to 10 for both the data stream and memory samples. To ensure a fair comparison, the same optimizer AdamW is applied for all experiments. Note that two separate AdamW optimizers are utilized for the two student models in our model. We tune the learning rate, weight decay, and other hyper-parameters for each baseline by maximizing its FAA score. Then, they are maintained when applying CCL-DC and our approach. Refer to the  YAML files in our code repository for their detailed values. The values of $\gamma$, $\Delta$ and $\alpha$ are assigned to 0.5, 1 Task and 0.01, respectively. The temperature hyper-parameters $\tau$ in $L_{\mathrm{KD}}$ and $L_{\mathrm{GKD}}$ are configured at $4.0$. The optimal value of $\lambda$ depends on the baseline, so refer to the released code for its exact configuration.

\Rt{In current implementation with two students, we adopt a deterministic and balanced configuration with $r^1=r^2=0.5$, which can be viewed as a special case of the Dirichlet-based formulation in Eq. (1). This decision is driven by stability concerns in OCIL, where training is tightly constrained and highly sensitive to any extra sources of randomness. Due to the different data augmentation strategies and independent optimizers, the two students still exhibit sufficient diversity, resulting in distinct optimization trajectories. Therefore, even with equal weights, the aggregation still effectively integrates complementary information while preserving stable training dynamics.}

Different augmentation techniques are applied to two student models. For student 1, we apply a transformation operation including random crop, random horizontal flip, color jitter and random grayscale. In contrast, RandAugment \cite{cubuk2020randaugment} is used for student 2. It contains two extra hyper-parameters $N$ and $M$, which are set as 3 and 15 in all experiments. Besides, several baselines possess their own data augmentation strategies. For a fair comparison, both models trained with CCL-DC and our method maintain these unique data augmentation techniques. Specifically, the ER, SCR, PCR and HPCR utilize random cropping, horizontal flipping, color jitter and random grayscale. The color jitter parameters are set to (0.4, 0.4, 0.4, 0.1) with a probability of $0.8$, while the random grayscale is applied with a probability of $0.2$. For ER-ACE, OCM, GSA and MOSE-MOE, the augmentation consists of random cropping and random horizontal flip. Notably, OCM introduces global rotation augmentation combined with inner flipping, generating $15$ times more training samples. GSA and MOSE-MOE adopt the inner flip operation to double the training samples.

\Rt{The proposed method is implemented using the PyTorch framework \cite{paszke2019pytorch}. All experiments were conducted on a server with 128 GB RAM and four NVIDIA GeForce RTX 3090 GPUs. However, the actual GPU requirements vary by datasets and methods. For CIFAR-100, all methods can be trained using a single GPU, whereas some methods needs multiple GPUs on Tiny-ImageNet and ImageNet-100.}

\begin{figure*}[!t]
    \centering     
    \includegraphics[width=\linewidth]{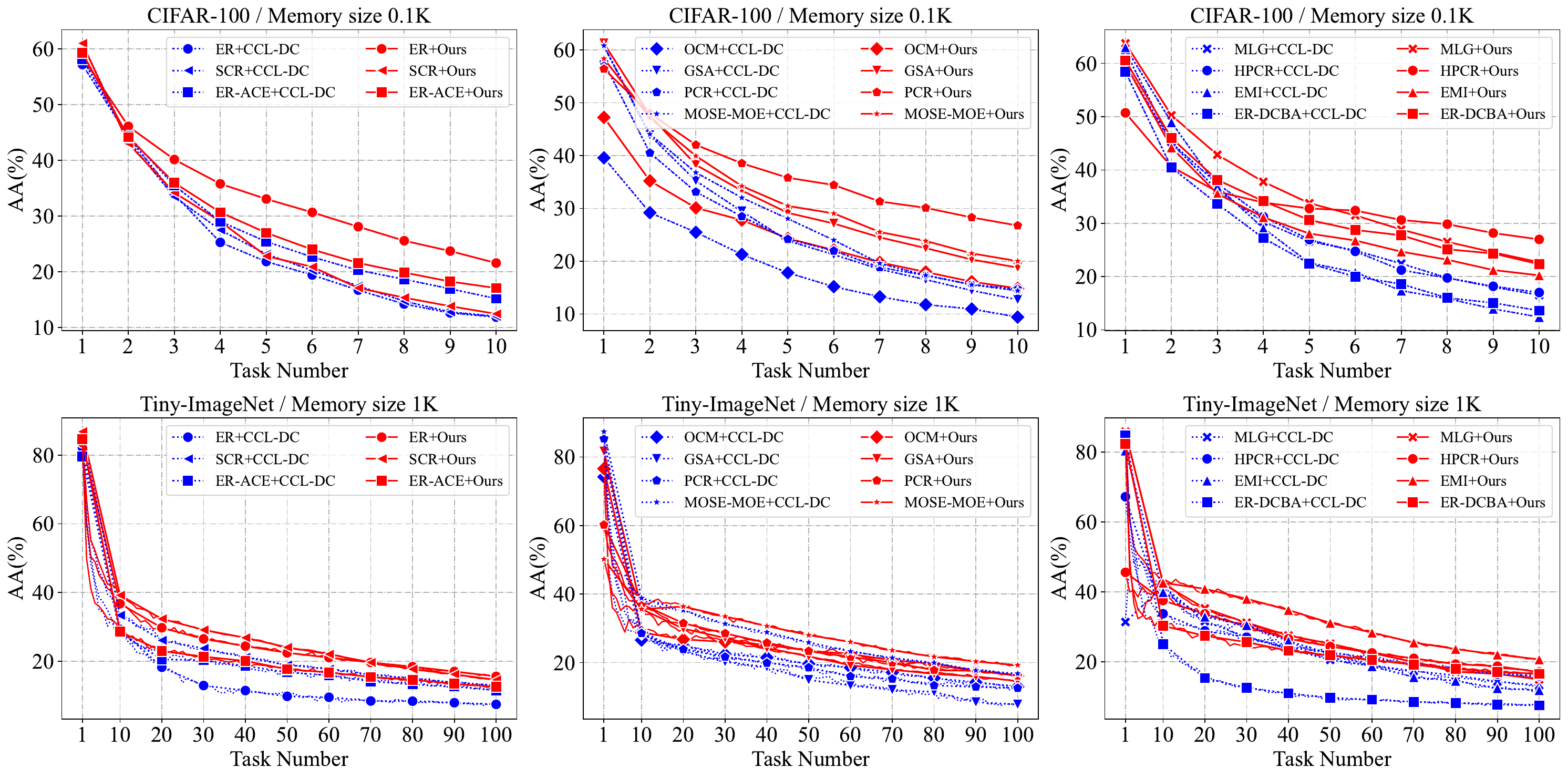}
    \caption{The average accuracy after each incremental step for various baselines when incorporating CCL-DC and ours on CIFAR-100 and Tiny-ImageNet. Diverse markers represent distinct baselines, where the colors blue and red denote \textcolor{blue}{CCL-DC} and \textcolor{red}{Ours}, respectively.}
    \label{incremental_line}
\end{figure*}

\begin{figure*}[t!]
    \centering    
    \includegraphics[width=\linewidth]{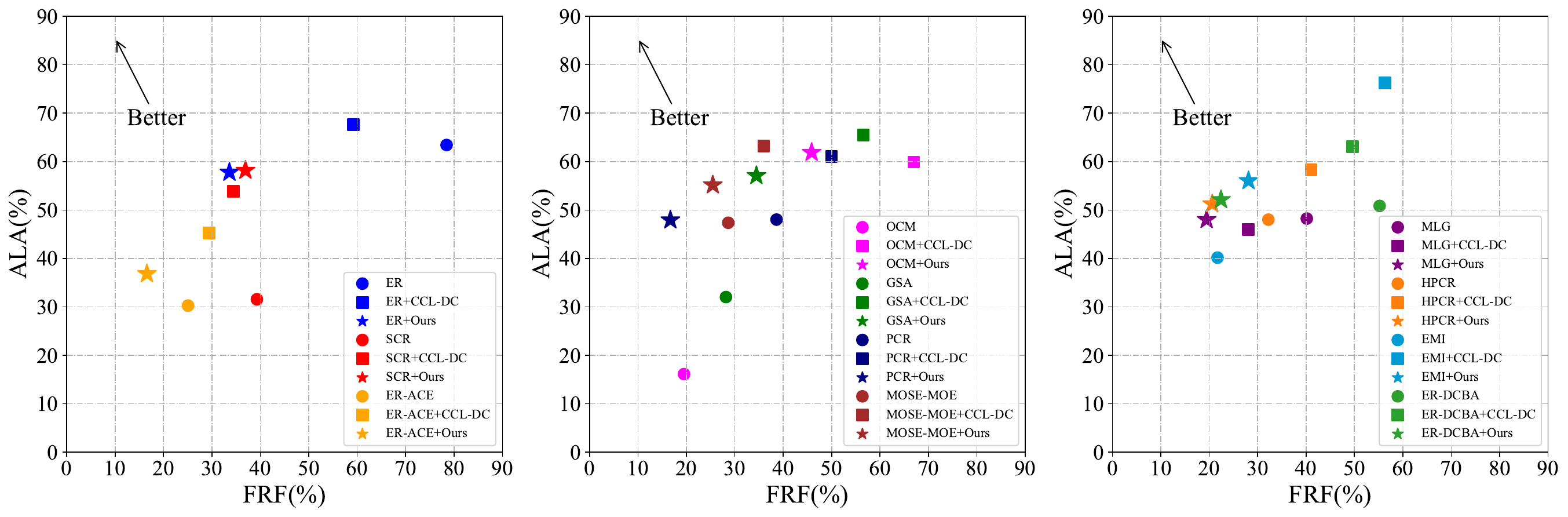}
    \caption{Comparison of stability and plasticity balance on CIFAR-100 at 1K memory. The {\Large $\bullet$} (circles), $\blacksquare$ (squares), and $\bigstar$ (stars) represent the baseline, CCL+DC, and our method, respectively. Various colors are used to distinguish the baselines. The closer to the left upper corner, the better the balance is struck.}
    \label{Stability and plasticity trade-off}
\end{figure*}

\subsection{Comparison with SOTA Methods}
\Rt{To evaluate the effectiveness of the proposed method, we applied it to several typical OCIL methods, including ER \cite{chaudhry2019tiny}, SCR \cite{mai2021supervised}, ER-ACE \cite{caccia2022new}, OCM \cite{guo2022online}, GSA \cite{guo2023dealing}, PCR \cite{lin2023pcr}, MOSE-MOE \cite{yan2024orchestrate}, MLG\cite{liang2025masking}, HPCR\cite{10839221}, EMI\cite{11360301}, and ER-DCBA\cite{11433804}. For comparative analysis, we additionally present the results of these baseline methods, and their combination with CCL-DC \cite{wang2024improving}.}

Table~\ref{restrict memory} and ~\ref{standard memory} present the FAA performance of all methods across the three datasets. Since we change the memory batch size from 64 to 10, the overall FAA scores are lower than those reported in the original papers. Nevertheless, in both tables, our method consistently enhances the performance of various baselines by a large margin, and surpasses the baselines+CCL-DC in nearly all cases. For example, when using ER as the baseline, we achieve an accuracy increase of $9.8\% \sim 12.9\%$ over CCL-DC on the CIFAR-100 dataset. When working with highly limited memory capacities, the CCL-DC method exhibits only slight improvement, even occasionally reduces performance for some latest methods. Conversely, our approach always enhances FAA significantly. \Rt{Using the latest ER-DCBA \cite{11433804} as an example, incorporating CCL-DC results in FAA changes of 0.0\%, 2.1\%, and 5.8\% under limited-memory settings on ImageNet-100. By comparison, our approach achieves substantially larger improvements of 7.2\%, 12.1\%, and 14.3\%. }

Fig. \ref{incremental_line} displays the average accuracy of all observed tasks after each incremental step on CIFAR-100 ($M_s$ = 0.1K) and Tiny-ImageNet ($M_s$=1K). The blue and red lines represent the accuracy curves after the baselines are combined with CCL-DC and our method, respectively. Obviously, for all baselines, our approach consistently attains the highest average accuracy at every step. 

To evaluate the balance between stability and plasticity, we visualize the interplay between ALA and FRF on CIFAR-100 at 1K memory in Fig.~\ref{Stability and plasticity trade-off}. The closer a point lies to the top-left corner, the better the balance is struck. In comparison with CCL-DC, our method achieves a lower forgetting rate while maintaining comparable learning accuracy. This demonstrates that our approach achieves a more favorable balance.

\setlength{\tabcolsep}{4pt}
\begin{table*}[ht]
    \caption{Effect of proposed modules when separately using ER, GSA and MLG as baseline. The terms ``KD", ``Fuse" and ``GKD" denote the KD between students, parameters fusion, and KD between GWM and each student, respectively.}
    \centering
    
    \resizebox{\textwidth}{!}{
        \begin{tabular}{c|ccc|ccc|ccc|ccc|ccc}
        \toprule
        \multicolumn{4}{c|}{Dataset} & \multicolumn{12}{c}{CIFAR-100} \\
        \midrule
        \multicolumn{4}{c|}{Memory Size($M_s$)} & \multicolumn{3}{c|}{0.1K} & \multicolumn{3}{c|}{0.5K} & \multicolumn{3}{c|}{1K} & \multicolumn{3}{c}{5K} \\
        \midrule
        Base. & KD & Fuse & GKD & FAA$\uparrow$ & FRF $\downarrow$ & ALA $\uparrow$ &FAA$\uparrow$ & FRF $\downarrow$ & ALA $\uparrow$ & FAA$\uparrow$ & FRF $\downarrow$ & ALA $\uparrow$ & FAA$\uparrow$ & FRF $\downarrow$ & ALA $\uparrow$\\
        \midrule
        \multirow{4}{*}{ER} 
        & & & & 7.4\pms{0.7} & 89.2\pms{0.6} & 64.0\pms{1.3}  & 10.9\pms{1.1} & 83.9\pms{1.6} & 64.0\pms{1.3} & 14.3\pms{1.0} & 78.5\pms{1.9} & 63.4\pms{1.3} & 29.2\pms{1.6} & 53.0\pms{3.3} & 61.6\pms{1.3}\\
        & \checkmark & & &  9.5\pms{0.8} & 87.3\pms{0.6} & 71.2\pms{1.1}  &  16.9\pms{1.3} & 77.4\pms{1.4} & 71.3\pms{1.3} & 23.6\pms{0.9} & 67.2\pms{1.4} & 69.8\pms{1.3} & 33.2\pms{1.0} & 52.3\pms{1.7} & 69.4\pms{1.7}\\
        & \checkmark & \checkmark & & 11.5\pms{0.6} & 85.1\pms{0.7} & 74.1\pms{1.1} & 21.4\pms{1.2} & 71.8\pms{1.7} & 73.3\pms{1.1} & 27.9\pms{0.8} & 62.4\pms{1.4} & 72.4\pms{0.7} & 36.5\pms{1.1} & 50.1\pms{1.3} & 72.7\pms{1.2}\\
        & \checkmark & \checkmark & \checkmark & 21.6\pms{1.4} & 66.1\pms{2.1} & 61.0\pms{1.5} & 34.5\pms{0.8} & 33.9\pms{1.8} & 52.1\pms{1.6} & 38.4\pms{0.9} & 33.6\pms{2.0} & 57.8\pms{1.4} & 43.7\pms{1.6} & 27.0\pms{3.9} & 59.9\pms{1.8}\\
        
        \midrule
        \multirow{4}{*}{GSA} 
        & & & & 12.1\pms{0.6} & 75.2\pms{2.1} & 44.8\pms{3.8} & 19.8\pms{1.8} & 42.8\pms{5.6} & 34.8\pms{2.4} & 23.8\pms{0.9} & 28.2\pms{3.0} & 32.0\pms{1.4} & 27.9\pms{1.9} & 19.6\pms{3.4} & 33.1\pms{1.4}\\
        & \checkmark & & & 12.3\pms{0.8} & 82.7\pms{1.9} & 66.1\pms{1.8} & 23.4\pms{0.8} & 62.8\pms{0.9} & 60.4\pms{1.3} & 31.5\pms{1.1} & 45.4\pms{3.2} & 57.0\pms{2.1} & 40.0\pms{1.8} & 28.5\pms{3.9} & 56.1\pms{2.0}\\
        & \checkmark & \checkmark & & 13.7\pms{0.9} & 81.1\pms{1.9} & 66.8\pms{1.8} & 26.1\pms{0.8} & 58.5\pms{1.2} & 60.9\pms{2.2} & 33.5\pms{1.3} & 43.3\pms{2.7} & 58.4\pms{2.1} & 41.9\pms{1.5} & 29.5\pms{2.4} & 59.6\pms{1.9}\\
        & \checkmark & \checkmark & \checkmark & 18.8\pms{1.3} & 70.3\pms{2.0} & 61.3\pms{3.5} & 32.0\pms{1.5} & 46.0\pms{0.9} & 58.5\pms{2.3} & 37.7\pms{0.9} & 34.5\pms{1.7} & 57.1\pms{1.4} & 45.8\pms{1.6} & 22.0\pms{3.0} & 58.8\pms{1.8}\\
        \midrule
        \multirow{4}{*}{MLG} 
        & & & & 15.6\pms{1.1} & 67.0\pms{3.0} & 47.7\pms{0.7} & 24.4\pms{0.8} & 50.9\pms{2.2} & 50.2\pms{0.9} & 28.5\pms{1.0} & 40.2\pms{2.5} & 48.2\pms{1.6} & 33.6\pms{0.8} & 28.0\pms{2.6} & 47.1\pms{1.2}\\
        & \checkmark & & & 17.5\pms{0.9} & 67.7\pms{3.0} & 56.0\pms{1.1} & 26.9\pms{1.1} & 49.5\pms{2.8} & 54.4\pms{0.8} & 30.7\pms{1.0} & 40.4\pms{4.0} & 52.5\pms{1.5} & 35.0\pms{0.9} & 27.8\pms{2.9} & 49.0\pms{0.9} \\
        & \checkmark & \checkmark & & 18.4\pms{1.1} & 64.9\pms{3.5} & 54.1\pms{0.8} & 28.1\pms{0.8} & 44.7\pms{2.2} & 52.0\pms{1.4} & 31.7\pms{0.8} & 35.2\pms{3.1} & 49.9\pms{1.6} & 36.1\pms{0.8} & 22.2\pms{3.4} & 47.0\pms{1.6} \\
        & \checkmark & \checkmark & \checkmark & 22.7\pms{0.8} & 59.2\pms{0.9} & 54.6\pms{0.7} & 34.6\pms{0.5} & 28.9\pms{1.4} & 49.3\pms{0.6} & 38.4\pms{1.1} & 19.5\pms{2.9} & 48.0\pms{0.9} & 43.5\pms{0.9} & 13.1\pms{3.4} & 49.6\pms{1.4} \\
        \midrule
        \multicolumn{4}{c|}{Dataset} & \multicolumn{12}{c}{ImageNet-100} \\
        \midrule
        \multicolumn{4}{c|}{Memory Size($M_s$)} & \multicolumn{3}{c|}{0.5K} & \multicolumn{3}{c|}{1K} & \multicolumn{3}{c|}{5K} & \multicolumn{3}{c}{10K}\\
        \midrule
        
        Base. & KD & Fuse & GKD & FAA$\uparrow$ & FRF $\downarrow$ & ALA $\uparrow$ & FAA$\uparrow$ & FRF $\downarrow$ & ALA $\uparrow$ & FAA$\uparrow$ & FRF $\downarrow$ & ALA $\uparrow$ & FAA$\uparrow$ & FRF $\downarrow$ & ALA $\uparrow$\\
        \midrule
        \multirow{4}{*}{ER} 
        & & & & 11.6\pms{1.6} & 77.8\pms{3.8} & 49.0\pms{2.1} & 14.9\pms{0.8} & 69.1\pms{3.0} & 47.2\pms{2.7} & 20.9\pms{2.2} & 64.9\pms{3.8} & 57.9\pms{1.2} & 24.5\pms{2.0} & 58.2\pms{4.3} & 58.5\pms{3.5}\\
        & \checkmark & & & 14.0\pms{0.7} & 82.1\pms{1.1} & 71.9\pms{1.0} & 20.5\pms{0.8} & 72.9\pms{1.0} & 71.6\pms{1.2} & 35.9\pms{1.3} & 49.4\pms{2.2} & 69.8\pms{1.6} & 37.6\pms{1.9} & 46.2\pms{3.4} & 69.7\pms{1.4}\\
        & \checkmark & \checkmark & & 17.3\pms{0.7} & 78.3\pms{1.1} & 75.0\pms{0.9} & 24.9\pms{0.9} & 67.9\pms{0.9} & 74.4\pms{1.9} & 39.1\pms{0.7} & 47.1\pms{0.9} & 73.1\pms{1.0} & 41.6\pms{1.2} & 43.6\pms{0.9} & 73.4\pms{1.5}\\
        & \checkmark & \checkmark & \checkmark & 21.9\pms{0.3} & 71.0\pms{0.7} & 72.2\pms{1.1} & 28.9\pms{0.7} & 60.5\pms{1.2} & 71.4\pms{1.1} & 40.6\pms{1.0} & 42.8\pms{1.3} & 70.8\pms{1.7} & 42.4\pms{1.3} & 40.4\pms{1.1} & 71.4\pms{1.7}\\

        \midrule
        \multirow{4}{*}{GSA} 
        & & & & 15.7\pms{1.6} & 69.0\pms{3.2} & 47.3\pms{2.2} & 20.2\pms{1.1} & 55.1\pms{2.3} & 43.0\pms{3.0} & 32.8\pms{2.1} & 13.6\pms{3.5} & 35.7\pms{1.5} & 35.9\pms{1.2} & 11.0\pms{1.1} & 36.4\pms{0.7}\\
        & \checkmark & & & 17.0\pms{2.0} & 76.9\pms{2.1} & 68.6\pms{2.1} & 24.8\pms{1.8} & 64.8\pms{3.1} & 67.2\pms{0.9} & 42.6\pms{1.1} & 30.2\pms{2.6} & 60.5\pms{1.5} & 46.1\pms{1.8} & 23.5\pms{2.5} & 59.9\pms{1.7}\\
        & \checkmark & \checkmark & & 18.7\pms{0.9} & 75.5\pms{1.0} & 71.4\pms{0.7} & 25.8\pms{0.8} & 64.2\pms{1.5} & 68.9\pms{1.1} & 43.9\pms{1.0} & 30.8\pms{2.0} & 63.0\pms{1.1} & 48.0\pms{0.7} & 24.5\pms{0.9} & 63.5\pms{0.7}\\
        & \checkmark & \checkmark & \checkmark & 23.3\pms{1.4} & 66.1\pms{1.9} & 65.8\pms{1.8} & 30.2\pms{3.2} & 53.5\pms{4.8} & 63.2\pms{1.6} & 45.9\pms{1.4} & 24.5\pms{1.5} & 60.6\pms{1.1} & 49.0\pms{1.3} & 20.4\pms{3.9} & 61.3\pms{1.3}\\

        \midrule
        \multirow{4}{*}{MLG} 
        & & & & 22.8\pms{0.8} & 53.3\pms{1.7} & 48.7\pms{1.3} & 27.6\pms{1.0} & 42.1\pms{2.6} & 47.8\pms{1.1} & 33.3\pms{1.6} & 24.8\pms{3.8} & 45.0\pms{1.2} & 34.6\pms{0.8} & 21.6\pms{4.4} & 44.5\pms{1.9} \\
        & \checkmark & & & 24.0\pms{0.9} & 59.6\pms{1.8} & 59.4\pms{1.3} & 27.8\pms{1.5} & 50.8\pms{3.8} & 56.7\pms{1.5} & 34.9\pms{1.2} & 31.1\pms{2.8} & 50.7\pms{1.2} & 34.8\pms{0.9} & 31.0\pms{1.7} & 50.3\pms{0.9} \\
        & \checkmark & \checkmark & & 25.8\pms{0.4} & 56.7\pms{1.4} & 59.5\pms{1.1} & 30.6\pms{0.4} & 46.1\pms{2.0} & 56.7\pms{1.1} & 36.8\pms{1.1} & 28.5\pms{2.0} & 51.6\pms{1.4} & 37.3\pms{0.2} & 27.2\pms{1.1} & 51.1\pms{1.1} \\
        & \checkmark & \checkmark & \checkmark & 30.9\pms{1.1} & 50.3\pms{3.1} & 62.0\pms{1.8} & 35.4\pms{0.7} & 41.9\pms{1.7} & 60.8\pms{1.3} & 43.5\pms{1.0} & 32.6\pms{1.6} & 64.3\pms{0.7} & 44.5\pms{0.9} & 31.7\pms{1.6} & 64.8\pms{1.1} \\
        \bottomrule
        \end{tabular}
    }
\label{Ablation Study}
\end{table*}
 
\subsection{Ablation Study}
\subsubsection{Effect of Proposed Modules}
Firstly, we analyze the effect of key modules including the KD between students (Eq. \eqref{eq:KD}), the fusion operation between GWM and students (Eq. \eqref{eq:fusion}), and the KD between GWM and each student (Eq. \eqref{eq:GWM-KD}), labeled as ``KD", ``Fuse", and ``GKD", respectively. \Rt{Table \ref{Ablation Study} reports the ablation study results on CIFAR-100 and ImageNet-100, where ER, GSA, and MLG are separately used as the baselines. For all baselines, incorporating any new modules can improve the FAA score. And combining all components achieves the best performance. Specifically, adding the KD loss between the two students yields a notable improvement in ALA. By aligning their predictions over various augmented views of the same data, it facilitates learning of new tasks. Moreover, the fusion operation generally increases stability and therefore lowers the FRF, as it guides the students into a more stable region. For GSA at some larger buffer sizes, the FRF remain steady even increases. We believe that the model has achieved a significantly lower FRF by retaining a larger number of samples. Meanwhile, the fusion also mitigates the impact of highly noisy samples, leading to more stable convergence on new tasks and consequently higher ALA. This benefit becomes even more evident on the more challenging ImageNet-100 dataset. Finally, incorporating GKD decreases the FRF in most cases, thereby further enhancing FAA.}

\subsubsection{Hyper-parameter Sensitivity}
\begin{figure*}[!t]
    \centering
    \begin{subfigure}{0.48\textwidth}
        \centering
        \includegraphics[width=\textwidth]{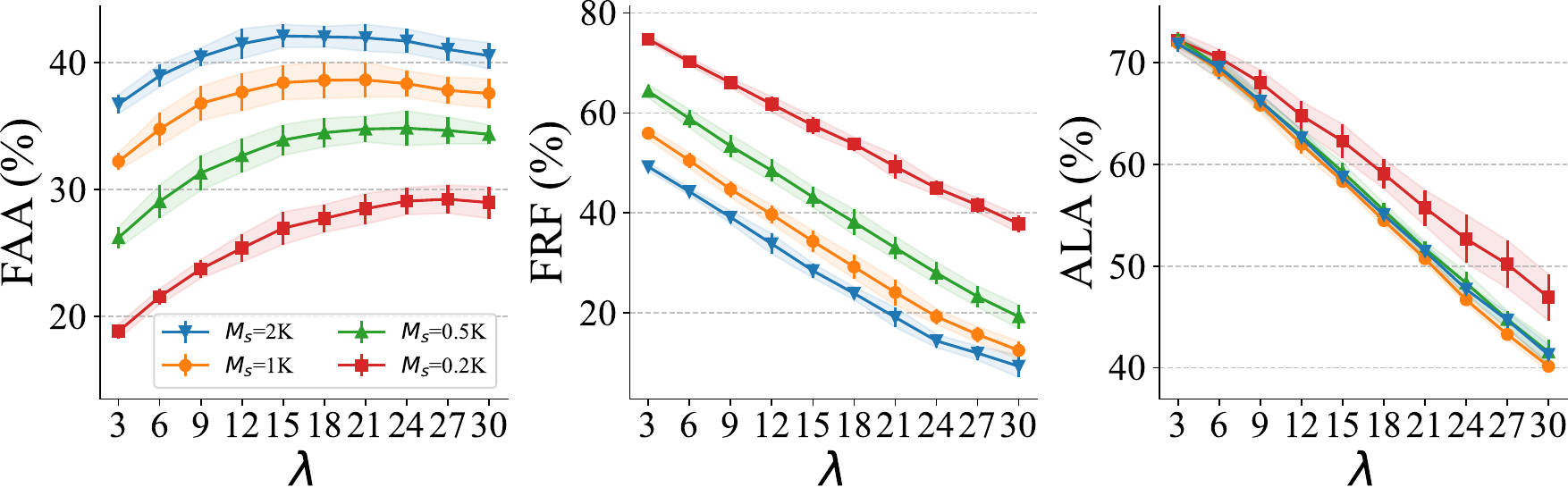}
        \caption{}
        \label{fig:lambda}
    \end{subfigure}
    \hfill
    \begin{subfigure}{0.48\textwidth}
        \centering
        \includegraphics[width=\textwidth]{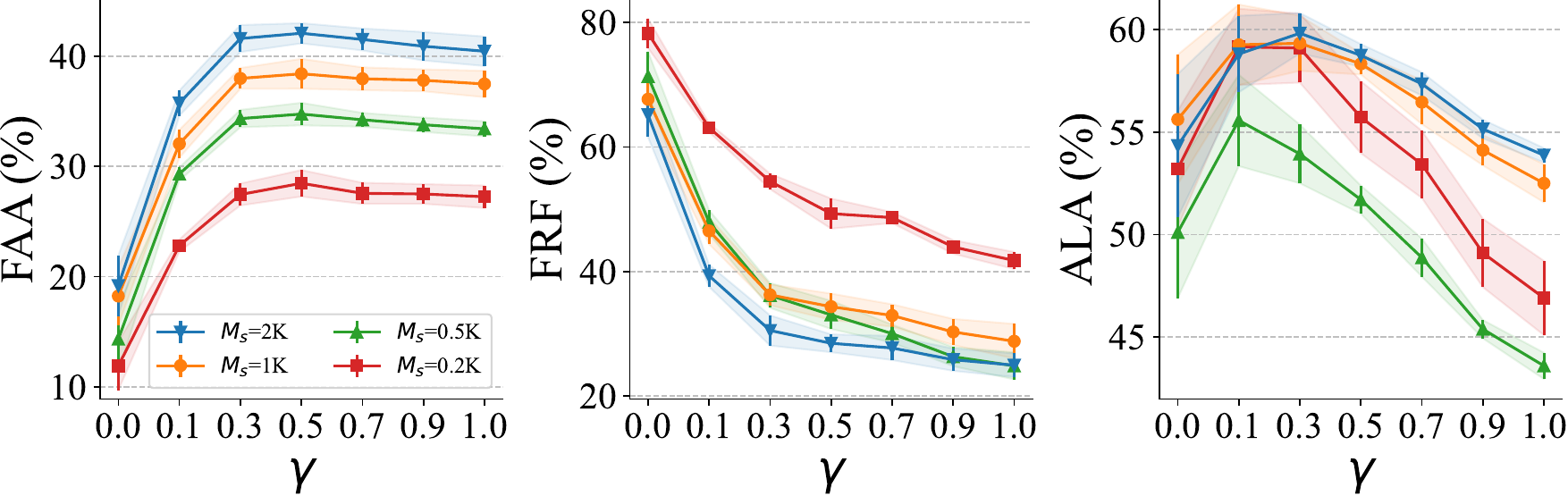}
        \caption{}
        \label{gamma}
    \end{subfigure}
    
    \begin{subfigure}{0.48\textwidth}
        \centering
        \includegraphics[width=\textwidth]{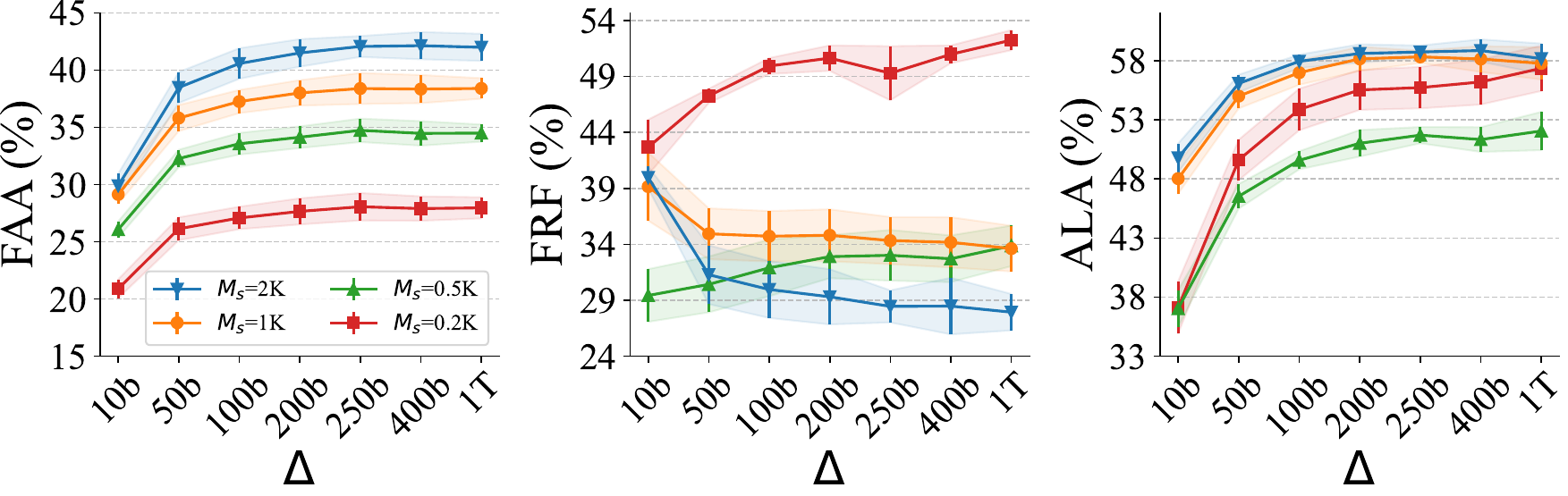}
        \caption{}
        \label{delta}
    \end{subfigure}
    \hfill
    \begin{subfigure}{0.48\textwidth}
        \centering
        \includegraphics[width=\textwidth]{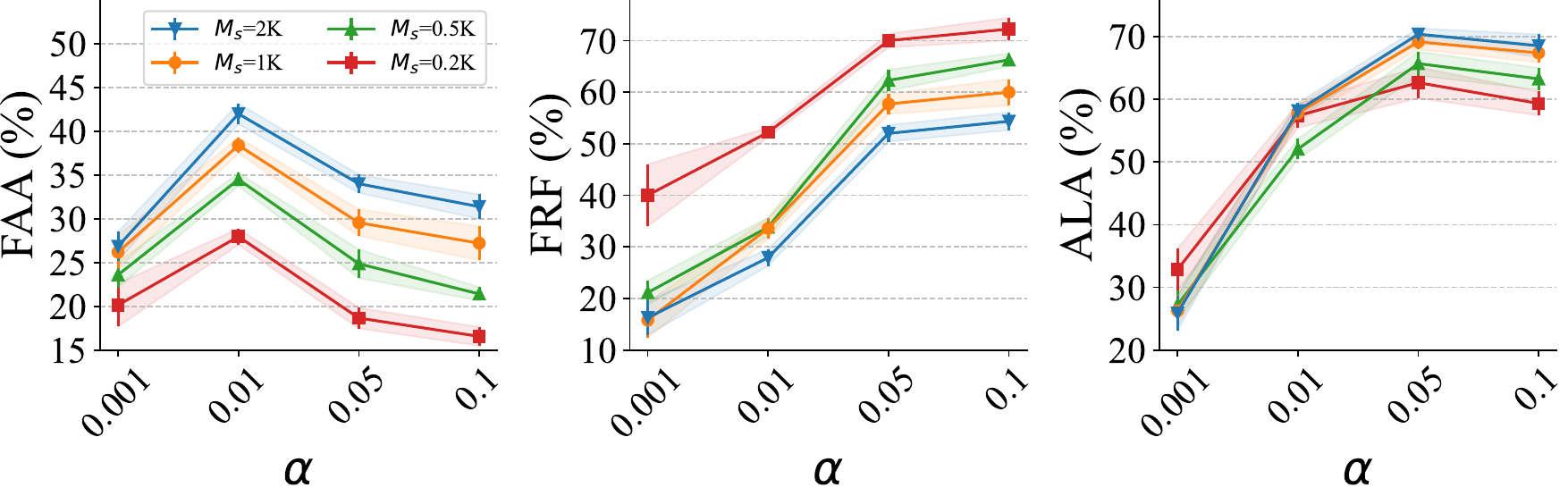}
        \caption{}
        \label{alpha}
    \end{subfigure}
    \caption{Impact of $\lambda$, $\gamma$, $\Delta$, and $\alpha$ in ER+Ours on CIFAR-100 with varying memory sizes. (a)Influence of $\lambda$ with $\gamma=0.5$, $\Delta=250$. (b)Influence of $\gamma$ with $\Delta=250$ and $\lambda$ searched in Fig (a). (c)Influence of $\Delta$ with $\gamma=0.5$ and $\lambda$ searched in Fig (a), where b and T in x-aixs are abbreviations for batch and task, respectively. (d)Influence of EMA coefficient $\alpha$.}
    \label{fig:parameter-sensitivity}
\end{figure*}

To illustrate the impact of the hyper-parameters $\lambda$, $\gamma$, $\Delta$ and $\alpha$, we have conducted experiments using ER+ours on CIFAR-100. The corresponding results are presented in Fig. \ref{fig:parameter-sensitivity}. We evaluated the effect of each hyper-parameter individually while keeping others constant.

Fig. \ref{fig:parameter-sensitivity} (a) shows the influence of loss coefficient $\lambda$ in Eq. \eqref{eq:okd}. When it increases, the FAA gradually ascends to a peak. After this point, the FAA remains relatively stable. This trend is consistent for various memory sizes. This indicates that moderate GKD between GWM and students is beneficial. It guides the students towards old flatter, more generalizable solutions, mitigating forgetting and improving overall FAA scores. Otherwise, excessively strong distillation causes students to blindly mimic the GWM, which restricts their ability to learn new tasks, ultimately resulting in a performance bottleneck. Furthermore, for smaller memory sizes, such as 0.2K and 0.5K, a larger $\lambda$ is required. This implies that when memory samples are much fewer, stronger regularization is necessary to assist the model in achieving optimal plasticity and stability balance.

\begin{figure*}[!t]
    \centering
    \includegraphics[width=0.9\linewidth]{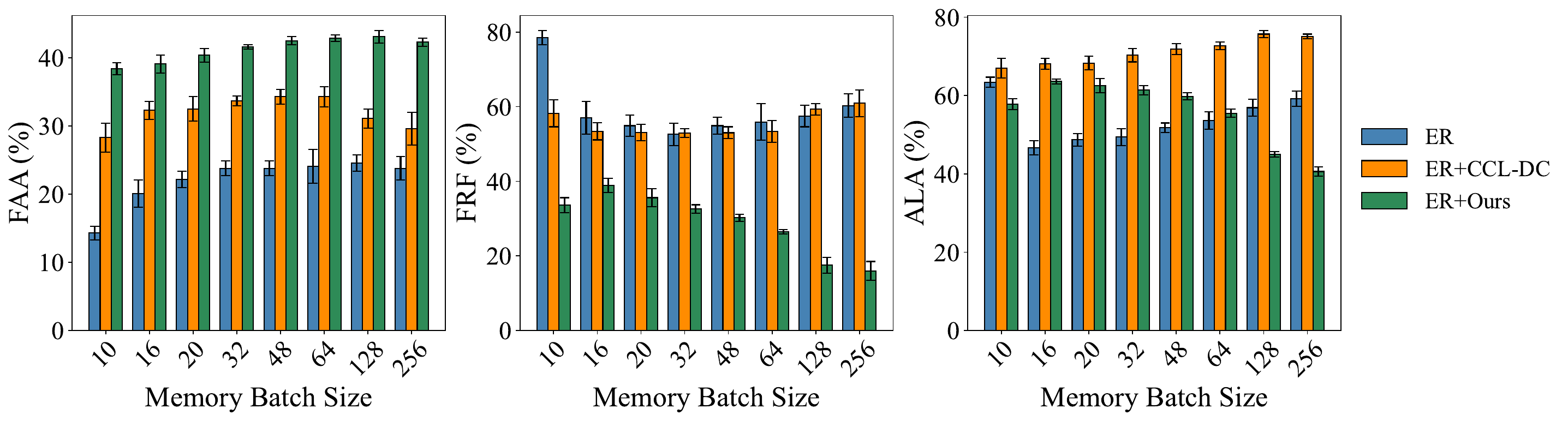}
    \caption{Influence of various memory batch size on the FAA, FRF, and ALA score with a fixed stream batch size of 10 on CIFAR-100 at 1K memory.}
    \label{fig:diff_mbs}
\end{figure*}
\begin{figure}[!t]
    \centering    \includegraphics[width=0.99\linewidth]{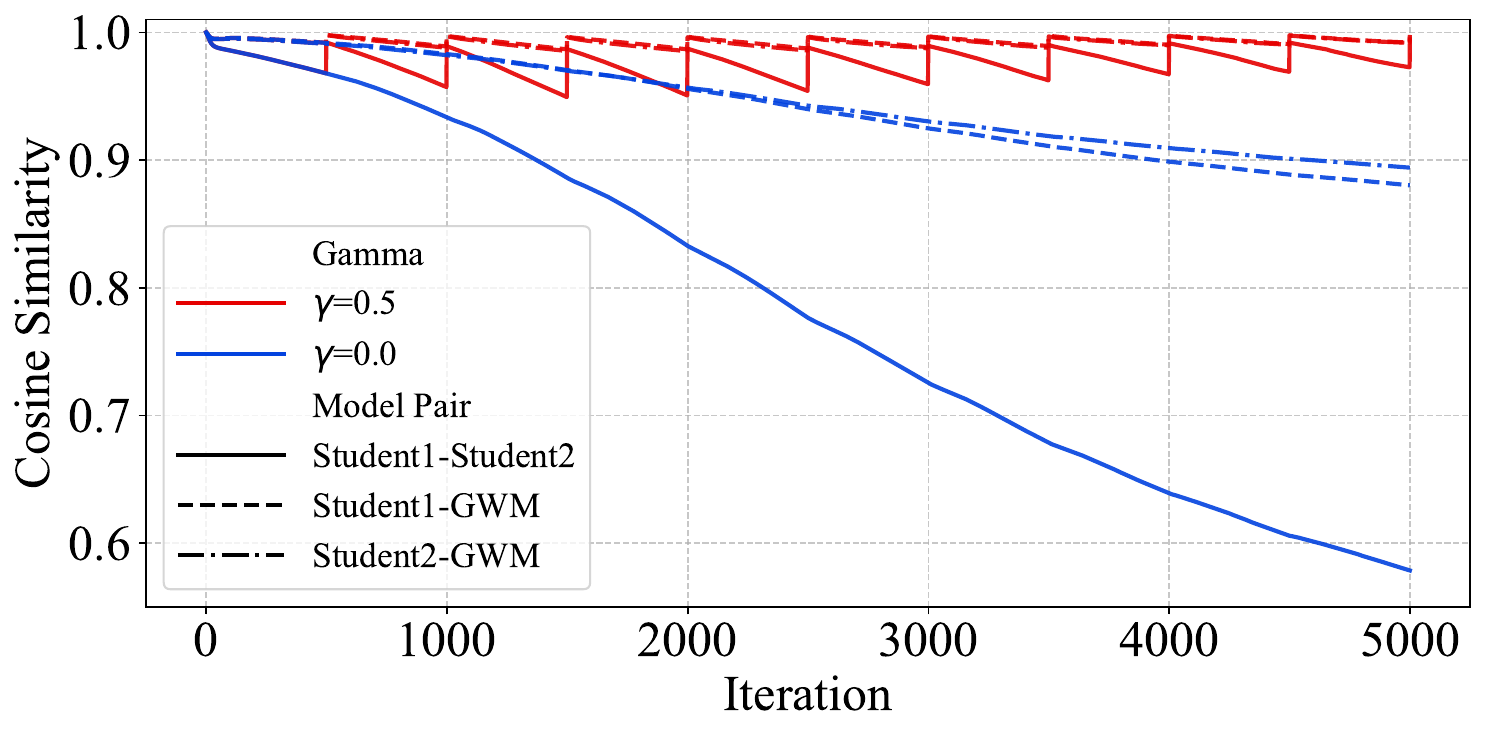}
    \caption{Comparison of parameter cosine similarity in ER+Ours concerning parameter fusion application between GWM and students on CIFAR-100 with 1K memory. Blue lines ($\gamma=0.0$) correspond to no parameter fusion while red lines ($\gamma=0.5$) represent an average fusion between GWM and students.}
    \label{fig:cosine similarity of gamma}
\end{figure}

In Eq. \eqref{eq:fusion}, we use $\gamma$ and $\Delta$ to denote the proportion of GWM and the interval of parameter fusion. Their influences are given in Fig. \ref{fig:parameter-sensitivity} (b) and (c). When $\gamma$ equals 0, the parameters of GWM are not fused back to students (equivalent to baseline + KD + GKD). In this case, the two students' parameters are only restricted by aligning their probability through KD and GKD. Without explicitly aligning the parameters of the two students, their parameter trajectories will progressively drift apart, leading to an ineffective construction of the GWM. Thus, the performance is the worst, reflecting the importance of our proposed fusion mechanism. As $\gamma$ increases, the FRF declines but the ALA firstly increases and then decreases rapidly. However, the overall FAA initially rises and then remains fairly constant once $\gamma$ exceeds 0.5. In our opinion, too strong fusion will suppress the model's adaptability to new tasks. Furthermore, Fig. \ref{fig:cosine similarity of gamma} visualizes the cosine similarity of parameters between GWM and each student model, along with the similarities between the two students. When parameter fusion back is not applied ($\gamma=0.0$), the two students gradually diverge over time. Conversely, this divergence is effectively reduced after employing parameter fusion mechanism. Since the fusion interval is set to 1 task, the $\gamma=0.5$ curve for CIFAR-100 with 10 tasks contains 10  peaks. These results indicate the importance of our fusion operation again.

\begin{table*}[!t]
    \caption{Comparison of the total running time (including both training time and inference time), GPU memory usage and FAA on CIFAR-100 at 1K memory.}
    \centering
    \begin{tabular}{c|ccc|ccc|ccc}
    \toprule
     Method & \multicolumn{3}{c|}{Baseline} & \multicolumn{3}{c|}{Baseline+CCL-DC\cite{wang2024improving}} & \multicolumn{3}{c}{Baseline+Ours} \\
    \midrule
     Metric & Time(s) & GPU(MB) & FAA(\%) & Time(s) & GPU(MB) & FAA(\%) & Time(s) & GPU(MB) & FAA(\%) \\
    \midrule
    ER\cite{chaudhry2019tiny}       & 181.27 & 296.99 & 14.3 & 638.10 & 1288.31 & 27.9 & 462.87 & 850.59 & 38.4  \\
    SCR\cite{mai2021supervised}      & 234.54 & 392.64 & 18.4 & 745.52 & 1473.45 & 35.3 & 641.48 &  1226.71 & 37.0 \\
    ER-ACE\cite{caccia2022new}   & 192.91 & 296.00 & 22.3 & 868.47 & 1289.06 & 31.3 & 602.29 & 987.37 & 34.0 \\
    OCM\cite{guo2022online}      & 934.50 & 2193.49 & 16.8 & 1907.11 & 4688.96 & 20.2 & 1845.09 & 4563.26 & 33.7  \\
    GSA\cite{guo2023dealing}      & 382.78 & 385.48 & 23.8 & 851.73 & 1092.97 & 31.5 & 772.58 & 977.76 & 37.7 \\
    PCR\cite{lin2023pcr}      & 305.70 & 377.02 & 29.3 & 779.52 & 1416.69 & 30.8 & 560.54  & 850.38 & 40.0 \\
    MOSE-MOE\cite{yan2024orchestrate} & 533.22 & 534.11 & 33.3 & 2297.89 & 1743.49 & 40.6 & 1985.36  & 1494.17 & 42.2\\
    MLG\cite{liang2025masking} & 371.13 & 398.67 &  28.5 & 1191.41 & 1616.60 & 32.5 & 809.73 & 1352.58 & 38.4  \\
    HPCR\cite{10839221} & 407.59 & 377.02 & 32.2 & 1170.19 & 1416.75& 34.5& 631.01 &  838.24 & 41.0\\
    EMI\cite{11360301} & 1937.33 & 4124.03 & 30.9 & 4396.79 & 8815.08 & 33.8 & 4379.66 & 8384.18 & 41.5  \\
    ER-DCBA\cite{11433804} & 461.17 & 415.76 & 23.3 & 1087.51 & 1495.35& 32.0 & 659.74&  1153.20 & 40.2\\
    \bottomrule
    \end{tabular}
\label{running time and gpu usage}
\end{table*}

As shown in Fig. \ref{fig:parameter-sensitivity} (c), too small fusion intervals, such as 10 batches, severely restrict the model's ability to explore new tasks, resulting in a considerably reduced ALA. Raising $\Delta$ facilitates the acquisition of new tasks, but it impacts stability for smaller memory. The two students are prone to overfitting to the limited old samples in the memory, which severely damages stability, as indicated by the FRF curves at 0.2K and 0.5K memory. When the interval is greater than 250 batches, the overall FAA remains relatively stable. We set $\Delta$ to 1 Task, obtaining the best performance across memory sizes. Meanwhile, this setup also lowers the fusion frequency, thereby reducing the computational budget.

Furthermore, Fig. \ref{fig:parameter-sensitivity} (d) shows the influence of the EMA coefficient $\alpha$ in Eq. \eqref{eq:GWM_EMA}. It can be observed that its value is vital in achieving an optimal balance between stability and plasticity. A smaller $\alpha$ results in a much slower update of the GWM, failing to fully capture new tasks. If $\alpha$ is too large, the GWM relies heavily on the current students, compromising stability. An $\alpha$ value of 0.01 strikes the best balance between preserving historical knowledge and adapting to new tasks.

\subsubsection{Influence of Memory Batch Size}
\Rt{To better reflect limited resources in practical scenarios, we set the memory batch size to 10, whereas most existing OCIL methods use 64. To analyze its influence, we have conducted experiments on CIFAR-100 using various batch sizes, whose results are shown in Fig. \ref{fig:diff_mbs}. Compared to both the ER and ER+CCL-DC, our approach still obtains the highest FAA when adopting larger memory batch sizes. Moreover, the FAA gain becomes even more pronounced once the memory batch size exceeds 64. In our opinion, the proposed GWM is more effective at mitigating the forgetting of old knowledge as the memory batch size grows, as reflected by the change in FRF.}

\subsubsection{Computational Cost}
Table \ref{running time and gpu usage} reports the total running time (including both training and inference) and GPU memory usage on CIFAR-100 at 1K memory. Both CCL-DC and our approach utilize dual learners, resulting in a higher computational expense than the baselines. However, compared to CCL-DC, we obtain much higher FAA accuracy with less running time and GPU memory consumption.

\subsubsection{Trade-off between Efficiency and Performance}
In previous sections, we merged the outputs of two students to produce the final prediction during inference. Although effective, this ensemble technique requires more computational overhead. To assess the ability of a single student, Table \ref{tab:average and individual comparison} gives the FAA achieved by an individual student as well as their average. Clearly, employing just one student slightly reduces accuracy. Nevertheless, when computational resources are extremely limited, deploying just one student is a feasible choice.

\Rt{One might ask whether employing more students could further improve performance. We conducted additional experiments on CIFAR-100 using more students, and the corresponding results are reported in Table \ref{tab:students_metrics}. When using 3 students, the FAA gain on CIFAR-100 is 0.1-1.4\% over various memory sizes, whereas GPU memory usage raises to 1216 MB from 850 MB. For 4 students, the GPU memory consumption grows further to 1583 MB, but the FAA performance stagnates or even degrades because the peers become more homogeneous. Therefore, we choose to use 2 students, providing a better balance between accuracy and efficiency.}

\begin{table}[!t]
    \caption{Comparison of FAA derived from the predicted probability of a single student and their average on CIFAR-100. S1, S2 denotes the first and second student, respectively.}
    \centering
    \resizebox{\linewidth}{!}{
    \setlength{\tabcolsep}{3pt}
    \begin{tabular}{c|c|cccccc}
        \toprule
         Method & $M_s$ & 0.1K & 0.2K &  0.5K & 1K & 2K & 5K\\
        \midrule
        \multirow{3}{*}{\parbox{1cm}{\centering ER+\\ Ours}} & S1 & 21.6\pms{1.3} & 27.7\pms{0.9} & 34.3\pms{1.0} & 38.3\pms{1.0} & 41.8\pms{1.3} & 43.6\pms{1.3}\\
         & S2 & 21.2\pms{1.5} & 27.6\pms{1.4} & 34.2\pms{0.8} & 38.1\pms{0.9} & 41.8\pms{1.3} & 43.7\pms{1.7}\\
         & Avg. & \textbf{21.6}\pms{1.4} & \textbf{28.0}\pms{0.9} & \textbf{34.5}\pms{0.8} & \textbf{38.4}\pms{0.9} & \textbf{42.0}\pms{1.2} & \textbf{43.7}\pms{1.6} \\
         \midrule
         \multirow{3}{*}{\parbox{1cm}{\centering GSA+\\ Ours}} & S1 & 18.7\pms{1.3} & 24.7\pms{1.2} & 31.6\pms{1.5} & 37.5\pms{0.9} & 42.5\pms{1.5} & 45.5\pms{0.8}\\
         & S2 & 18.8\pms{1.4} & 24.7\pms{1.4} & 31.9\pms{1.5} & 37.6\pms{0.8} & 42.5\pms{1.6} & 45.7\pms{0.5}\\
         & Avg. & \textbf{18.8}\pms{1.3} & \textbf{24.8}\pms{1.3} & \textbf{32.0}\pms{1.5} & \textbf{37.7}\pms{0.9} & \textbf{42.7}\pms{1.4} & \textbf{45.8}\pms{1.6}\\
         \bottomrule
    \end{tabular}
    }
    \label{tab:average and individual comparison}
\end{table}

\begin{table}[!t]
\centering
\caption{Influence of the number of student peers in ER+Ours on the performance of CIFAR-100 under different memory sizes. SN denotes the student number.}
\label{tab:students_metrics}
    \begin{tabular}{ccccccc}
    \toprule
    \multirow{2}{*}{SN} & \multicolumn{6}{c}{$M_s$} \\
    \cmidrule(lr){2-7}
     & 0.1K & 0.2K & 0.5K & 1K & 2K & 5K \\
    \midrule
    
    \multicolumn{7}{c}{\textbf{FAA (\%) $\uparrow$}} \\
    \midrule
    2 & $21.6\pms{1.4}$ & $28.0\pms{0.9}$ & $34.5\pms{0.8}$ & $38.4\pms{0.9}$ & $42.0\pms{1.2}$ & $43.7\pms{1.6}$ \\
    3 & $23.0\pms{0.8}$ & $28.1\pms{1.5}$ & $34.7\pms{0.6}$ & $39.2\pms{1.2}$ & $43.0\pms{1.4}$ & $44.4\pms{2.4}$ \\
    4 & $22.2\pms{0.9}$ & $28.0\pms{1.0}$ & $34.6\pms{0.5}$ & $38.4\pms{0.5}$ & $41.8\pms{0.7}$ & $42.4\pms{1.1}$ \\
    \midrule
    
    \multicolumn{7}{c}{\textbf{FRF (\%)} $\downarrow$} \\
    \midrule
    2 & $66.1\pms{2.1}$ & $52.2\pms{0.9}$ & $33.9\pms{1.8}$ & $33.6\pms{2.0}$ & $27.9\pms{1.6}$ & $27.0\pms{3.9}$ \\
    3 & $64.8\pms{1.5}$ & $53.0\pms{2.7}$ & $34.5\pms{1.6}$ & $32.6\pms{1.7}$ & $26.4\pms{2.2}$ & $25.9\pms{3.7}$ \\
    4 & $62.3\pms{2.3}$ & $47.5\pms{1.5}$ & $29.2\pms{2.4}$ & $29.9\pms{1.5}$ & $25.5\pms{2.5}$ & $26.4\pms{2.9}$ \\
    \midrule
    
    \multicolumn{7}{c}{\textbf{ALA (\%)} $\uparrow$} \\
    \midrule
    2  & $61.0\pms{1.5}$ & $57.4\pms{1.9}$ & $52.1\pms{1.6}$ & $57.8\pms{1.4}$ & $58.2\pms{1.3}$ & $59.9\pms{1.8}$ \\
    3  & $62.6\pms{1.0}$ & $58.9\pms{1.6}$ & $53.2\pms{1.1}$ & $58.0\pms{0.8}$ & $58.6\pms{1.1}$ & $60.0\pms{0.5}$ \\
    4 & $56.6\pms{1.4}$ & $53.0\pms{0.5}$ & $48.3\pms{1.6}$ & $54.9\pms{1.3}$ & $55.8\pms{1.1}$ & $57.2\pms{1.3}$ \\
    \bottomrule
    \end{tabular}
\end{table}

\subsubsection{Effect using Pre-trained Model}
To ensure a fair comparison, we employ the ResNet-18 without pre-training as our backbone. To assess proposed framework under pre-trained models, we further conduct experiments by substituting ResNet-18 with a pre-trained model. While the ViT model pre-trained on ImageNet-21k or ImageNet-1k is widely used, its supervised training lead to potential information leakage due to class overlap with datasets used in CIL \cite{momeni2025continual}. Therefore, we adopt the self-supervised DINOv2 model as the pre-trained backbone. Under this setting, we re-implement ER, ER+CCL-DC, and ER+Ours. All experimental configurations are identical to those used with ResNet-18, except that the learning rate is adjusted to 1e-05. The experimental results on CIFAR-100 and ImageNet-100 are shown in Table  \ref{tab:dinov2}. Owing to its stronger feature extraction capacity, leveraging a pre-trained model substantially improves the overall performance of all approaches. Even so, ER+Ours consistently achieves the highest FAA across all memory sizes, underscoring its effectiveness.

\begin{table*}[!t]
\centering
\caption{Comparison of FAA (\%) on CIFAR-100 and ImageNet-100 using ResNet-18 trained from scratch and a self-supervised pre-trained DINOv2 model under different memory sizes.}
    \centering
    \resizebox{\linewidth}{!}{
    \begin{tabular}{cc|cccccc|cccccc}
        \toprule
        \multirow{2}{*}{Method}& \multirow{2}{*}{Backbone} & \multicolumn{6}{c|}{CIFAR-100} & \multicolumn{6}{c}{ImageNet-100 }\\
        
        \cmidrule(lr){3-8} \cmidrule(lr){9-14}
         &  & 0.1K & 0.2K & 0.5K & 1K & 2K & 5K 
         & 0.2K &  0.5K &  1K & 2K & 5K & 10K
         \\
        \midrule
        \multirow{2}{*}{ER} & ResNet-18 & ~7.4\pms{0.7} & ~8.6\pms{0.5} & 10.9\pms{1.1} & 14.3\pms{1.0} & 20.3\pms{0.8} & 29.2\pms{1.6}

        &~8.1\pms{1.6} & 11.6\pms{1.6} & 14.9\pms{0.8} & 14.6\pms{2.0} & 20.9\pms{2.2} & 24.5\pms{2.0}
        
        \\
         & DINOv2 & 39.8\pms{2.3} & 49.7\pms{1.5} & 62.4\pms{3.1} & 71.5\pms{1.4} & 77.5\pms{1.7} & 82.3\pms{1.7} 
         
         & 54.2\pms{3.2} & 65.0\pms{2.7} & 72.6\pms{3.9} & 78.8\pms{1.4} & 85.7\pms{1.5} & 88.0\pms{0.7}
         \\
         
        \midrule
        \multirow{2}{*}{ER+CCL-DC}& ResNet-18 & 11.8\pms{1.1} & 15.1\pms{1.2} & 23.3\pms{1.4} & 27.9\pms{1.1} & 30.6\pms{2.0} & 31.2\pms{1.4} 
        
        & 11.8\pms{1.3} & 17.5\pms{1.1} & 24.3\pms{0.5} & 32.1\pms{1.7} & 37.4\pms{1.1} & 40.1\pms{0.4}
        \\
         & DINOv2 & 33.0\pms{1.6} & 47.7\pms{6.0} & 64.1\pms{2.5} & 74.5\pms{0.9} & 79.8\pms{2.2} & 84.7\pms{1.0} 

         & 55.5\pms{3.4} & 71.4\pms{3.1} & 79.9\pms{1.5} & 85.3\pms{0.6} & 90.2\pms{0.9} & 91.7\pms{0.9}
         \\
        \midrule
        
        \multirow{2}{*}{ER+Ours} & ResNet-18 & 21.6\pms{1.4} & 28.0\pms{0.9} & 34.5\pms{0.8} & 38.4\pms{0.9} & 42.0\pms{1.2} & 43.7\pms{1.6} 
        
        & 14.8\pms{1.1} & 21.9\pms{0.3} & 28.9\pms{0.7} & 32.9\pms{2.3} & 40.6\pms{1.0} & 42.4\pms{1.3}
        \\
         & DINOv2 & \textbf{67.2}\pms{1.5} & \textbf{76.3}\pms{0.9} & \textbf{85.0}\pms{0.7} & \textbf{87.5}\pms{1.1} & \textbf{89.4}\pms{0.7} & \textbf{91.0}\pms{0.3} 
         
         & \textbf{71.9}\pms{0.8} & \textbf{81.9}\pms{0.8} & \textbf{87.1}\pms{0.8} & \textbf{90.5}\pms{1.0} & \textbf{92.3}\pms{0.4} & \textbf{93.2}\pms{0.4}
         \\
        \bottomrule
    \end{tabular}
    }
\label{tab:dinov2}
\end{table*}

\subsubsection{Performance on Video-based Incremental Action Recognition}
To show the effectiveness beyond image recognition task, we further extend our framework to online class-incremental learning for video-based action recognition. Specifically, we adopt the Temporal Segment Networks (TSN) \cite{wang2018temporal}, which utilizes a ResNet-34 network pre-trained on ImageNet as the feature extractor. \Rt{The experimental results on UCF-101 \cite{soomro2012ucf101} with 10-tasks and 20-tasks splits are presented in Table \ref{tab:video_results}, where the memory size is set to 2020 videos.} Obviously, our method attains the highest FAA score under both 10-task and 20-task settings. Compared with ER+CCL-DC, it preserves a comparable ALA while markedly lowering the forgetting rate. These results demonstrate that our method can be effectively applied to video-based incremental action recognition task, highlighting its wide applicability.

\setlength{\tabcolsep}{3pt}
\begin{table}[!t]
\centering
\caption{Performance comparison on online video-based incremental action recognition on UCF-101.}
    \centering
    \resizebox{\linewidth}{!}{
    \begin{tabular}{cccc|ccc}
        \toprule
        Dataset & \multicolumn{3}{c|}{UCF-101 (10 tasks)} & \multicolumn{3}{c}{UCF-101 (20 tasks)} \\
        \midrule
        Metrics & FAA$\uparrow$ & FRF$\downarrow$ & ALA $\uparrow$ & FAA$\uparrow$ & FRF$\downarrow$ & ALA$\uparrow$\\
        \midrule
        ER & 43.9\pms{3.1} & 39.2\pms{5.0} & 71.0\pms{2.4} & 13.1\pms{3.1} & 72.8\pms{8.7} & 42.5\pms{2.6}\\
        ER+CCL-DC  & 51.2\pms{2.7} & 36.7\pms{1.6} & 80.3\pms{3.5} & 39.8\pms{2.7} & 51.9\pms{1.6} & 81.9\pms{3.7} \\
        \rowcolor{blue!10} 
        ER+Ours & \textbf{58.5}\pms{1.4} & \textbf{29.7}\pms{1.5} & \textbf{82.0}\pms{1.2} & \textbf{51.8}\pms{1.0} & \textbf{38.9}\pms{1.6} & \textbf{82.7}\pms{1.3} \\
        \bottomrule
    \end{tabular}
    }
\label{tab:video_results}
\end{table}

\subsection{Visualization}
\subsubsection{Analysis of Feature Drift}
\begin{figure}[!t]
    \centering    
    \includegraphics[width=\linewidth]{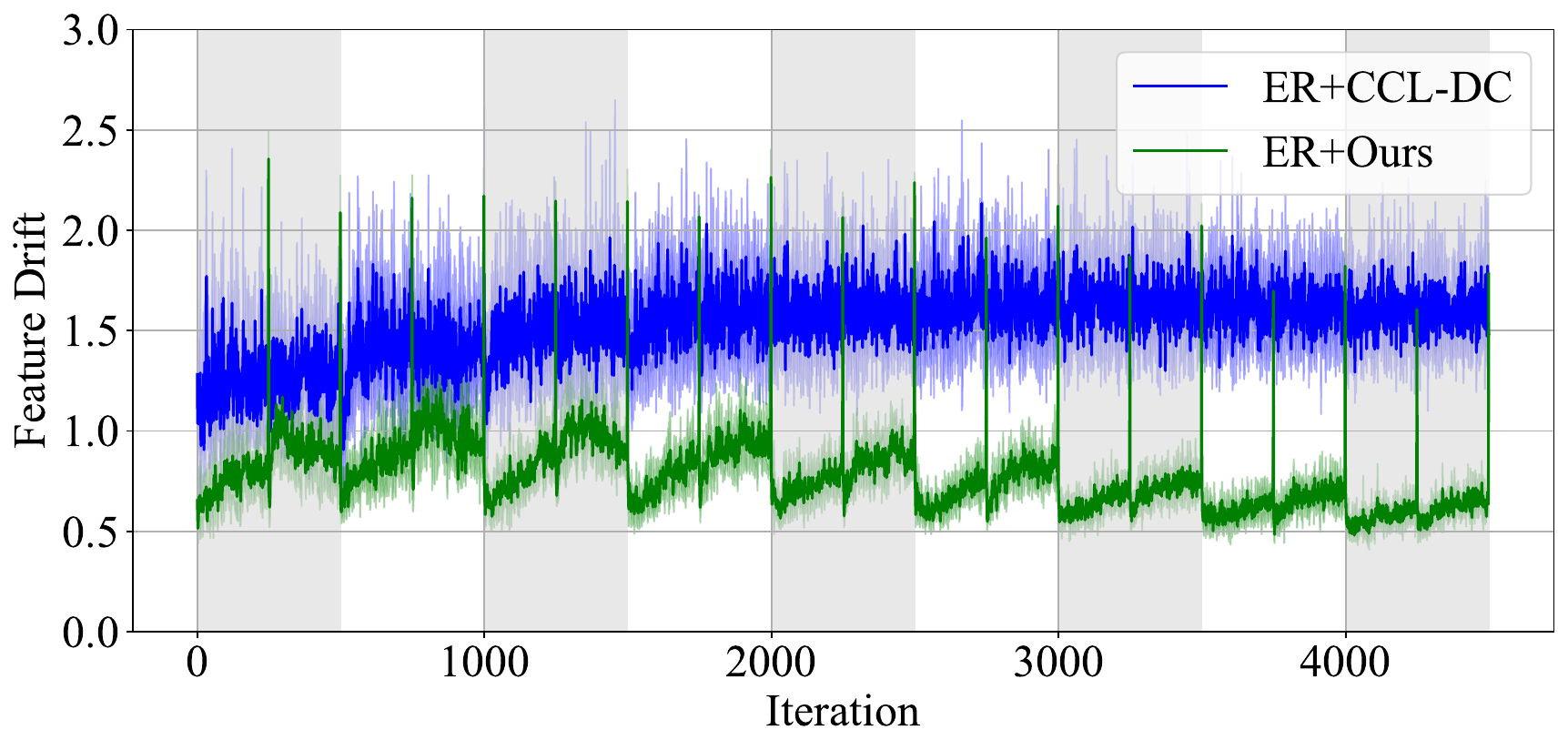}
    \caption{Comparison of feature drift of ER+CCL-DC and ER+Ours on CIFAR-100 at 0.5K memory.}
    \label{fig:drift}
\end{figure}

Due to feature conflict, the adjustment of parameters for acquiring a new task will cause an excessive shift in the features for previously learned tasks. To measure this feature drift, we employ the feature distance proposed in \cite{caccia2022new,michel2024rethinking}. Specifically, after the $i$-th iteration, we compute $\vert\vert f(\mathcal{X}_{old};\Theta_i)-f(\mathcal{X}_{old};\Theta_{i-1}) \vert\vert_2$, where $\mathcal{X}_{old}$ stands for memory images of old classes. Fig. \ref{fig:drift} visualizes the feature distance curves of ER+CCL-DC and ER+ours throughout the entire training process. It is evident that our approach significantly minimizes the feature drift, resulting in a smoother curve.

\begin{figure}[!t]
    \centering
    \includegraphics[width=\linewidth]{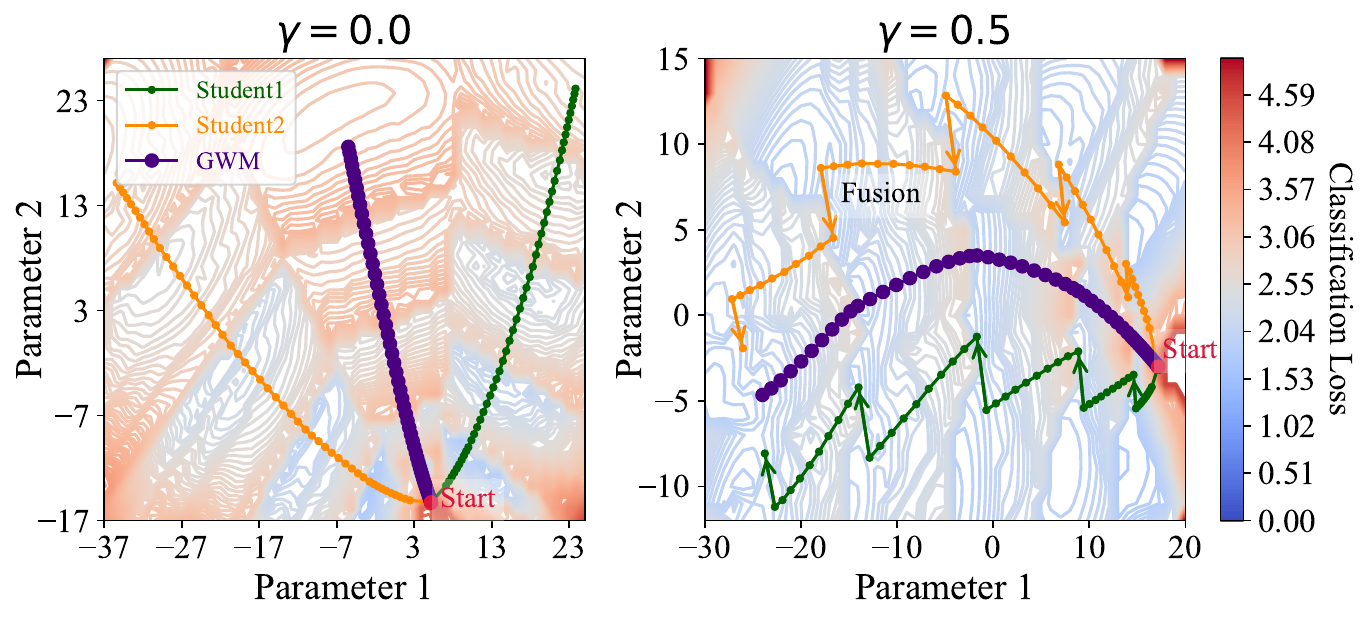}
    \caption{Visualization of loss trajectories under different fusion strengths $\gamma$ on CIFAR-100 with 1K memory.
    The background contours denote the classification loss projected onto a 2D parameter subspace via PCA. The loss trajectories for Student 1, Student 2, and GWM are shown as green, orange, and purple curves, respectively.
    \textbf{Left ($\gamma=0.0$)}: Without fusion, the two students gradually drift apart and become separated by high-loss barriers, leading to an ineffective GWM. \textbf{Right ($\gamma=0.5$)}: With periodic fusion, both students are repeatedly pulled toward the GWM, keeping all models within a common low-loss basin. This stabilizes optimization paths and enables effective parameter fusion, thereby improving stability.}
    
    \label{fig:landscape_gamma_0.0_0.5}
\end{figure}

\subsubsection{Visualization of Loss Trajectory}
Fig. \ref{fig:landscape_gamma_0.0_0.5} visualizes the loss trajectories of students and GWM for ER+Ours on CIFAR-100 with 1K memory size. Without fusion ($\gamma=0$), the parameters of the two students gradually drift apart since the KD loss just can restrict the output logits. Thus, as more tasks comes, the two students become separated by a high-loss region and fall into different basins of the loss landscape. This leads to a fail construction of GWM. In contrast, introducing periodic fusion ($\gamma=0.5$) fundamentally changes the optimization dynamics, which continuously pulls students' parameters back to a shared, stable anchor. This contraction mechanism prevents basin separation, preserves low-loss connectivity between students, and explains why periodic fusion guarantees the successful construction of GWM, ultimately enhancing stability.

\label{Section4:Experiments}
\section{Conclusion}
\label{Section5:Conclusion}
In this study, we introduced a novel cognitively-inspired framework for OCIL by instantiating the Global Workspace Theory (GWT). Specifically, we proposed a Global Workspace Model (GWM) that embodies the integrated knowledge of multiple student models via linear combination. We further design a broadcast mechanism, implemented as periodic parameter fusion, which acts as a contraction mapping to stabilize student optimization and maintain Linear Mode Connectivity. Coupled with a Multi-level Collaborative Distillation loss, our framework provides a principled solution to the stability-plasticity dilemma. Extensive experiments on three popular OCIL benchmarks demonstrate the effectiveness of our method in enhancing stability while maintaining plasticity, resulting in notable improvements in overall performance, especially for much smaller memory sizes. \Rt{In the future, we plan to explore a larger set of heterogeneous collaborative learners to better simulate the competition-coordination dynamics of GWT. Additionally, we intend to investigate adaptive competition mechanisms (e.g., data- or loss-dependent), as well as more powerful data augmentation techniques. By expanding the diversity of student inputs, we expect to further improve learning robustness and generalization.}


 
%

\bibliography{IEEEabrv,mybibfile}
\bibliographystyle{IEEEtran}


\begin{IEEEbiography}[{\includegraphics[width=1in,height=1.25in,clip,keepaspectratio]{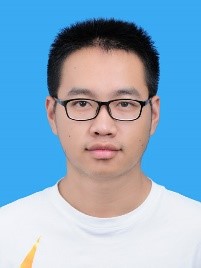}}]{Shibin Su} is currently a Master student in School of Computer Science, Northwestern Polytechnical University, Xi’an, China. His research interests include deep learning and continual learning.
\end{IEEEbiography}

\begin{IEEEbiography}[{\includegraphics[width=1in,height=1.25in,clip,keepaspectratio]{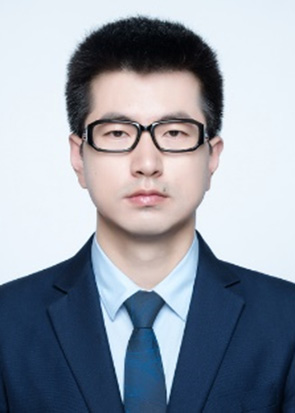}}]{Guoqiang Liang} received the B.S. in automation and the Ph.D. degrees in pattern recognition and intelligent systems from Xi’an Jiaotong University, Xi’an, China in 2012 and 2018 respectively. From Aug. 2018 to Aug 2020, he did the Post-Doctoral Research at the School of Computer Science, Northwestern Polytechnical University (NWPU), Xi’an, China. Currently, he is an associate professor at NWPU. His research interests include pattern recognition, machine learning and image analysis.
\end{IEEEbiography}

\begin{IEEEbiography}[{\includegraphics[width=1in,height=1.25in,clip,keepaspectratio]{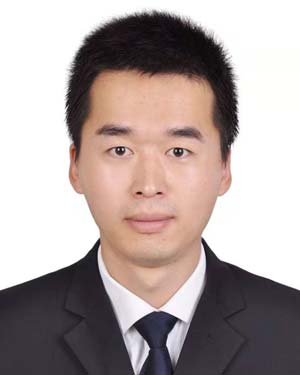}}]{De Cheng} received the B.S. and Ph.D. degrees from Xi’an Jiaotong University, Xi’an, China, in 2011
and 2017, respectively. He is currently an Associate Professor with the School of Telecommunications
Engineering, Xidian University, Xi’an. From 2015 to 2017, he was a Visiting Scholar with Carnegie
Mellon University, Pittsburgh, PA, USA. His research interests include pattern recognition, machine learning, and multimedia analysis.
\end{IEEEbiography}

\begin{IEEEbiography}[{\includegraphics[width=1in,height=1.25in,clip,keepaspectratio]{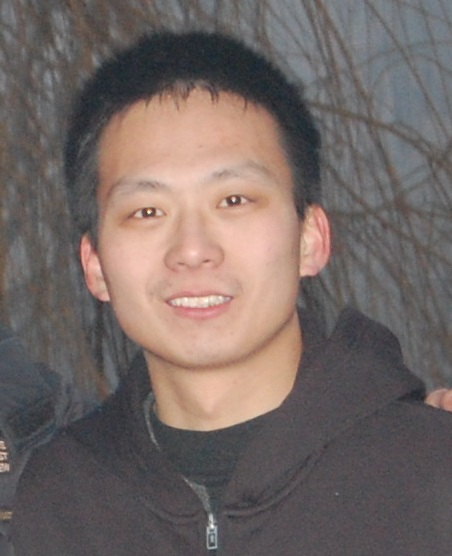}}]{Shizhou Zhang} received a B.E. and Ph.D. degree from Xi'an Jiaotong University, Xi'an, China, in 2010 and 2017, respectively. Currently, he is with Northwestern Polytechnical University as an associate professor. His research interests include content-based image analysis, pattern recognition and machine learning, specifically in the areas of deep learning based vision tasks such as image classification, object detection, re-identification and semantic parsing.
\end{IEEEbiography}

\begin{IEEEbiography}
[{\includegraphics[width=1in,height=1.25in,clip,keepaspectratio]{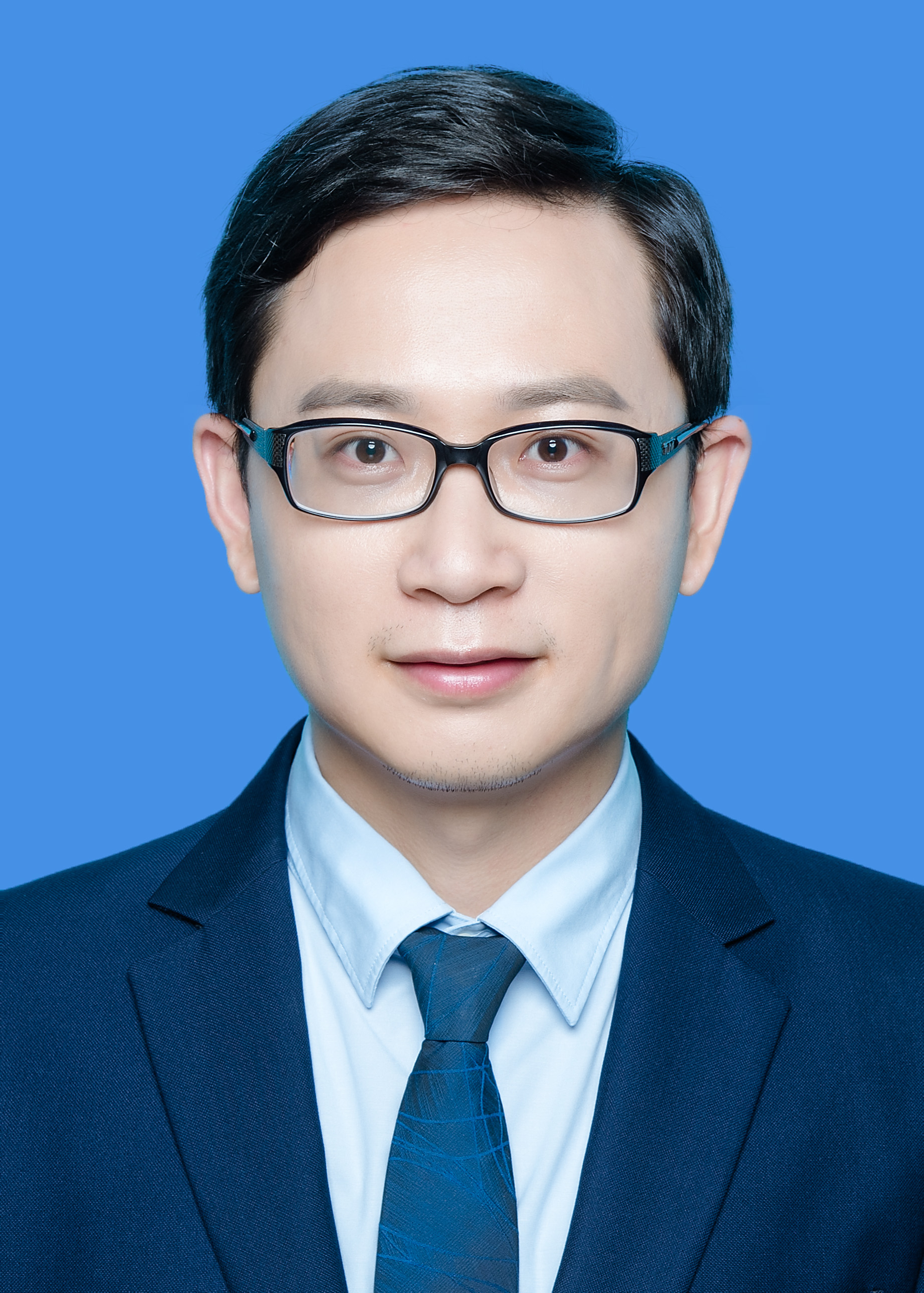}}]
{Lingyan Ran} received his B.S. and Ph.D. degrees from Northwestern Polytechnical University (NWPU), Xi'an, China, in 2011 and 2018. Earlier, he was a visiting scholar at Stevens Institute of Technology, Hoboken, NJ, from 2013 to 2015. He is currently an Associate Professor with the School of Computer Science, NWPU. His research interests include image classification, semantic segmentation, and change detection. He is currently a member of the China Computer Federation and China Society of Image and Graphics.
\end{IEEEbiography}

\vfill

\end{document}